%% file: sketchgen.tex
\documentclass[runningheads]{llncs}

\pdfoutput=1

\input{sections/preamble}

\usepackage{graphicx}
\usepackage{amsmath,amssymb} % define this before the line numbering.
\usepackage{color}
\begin{document}
\title{RNN-based Generative Model for Fine-Grained Sketching}

\author{
    Andrin Jenal
    \and
    Nikolay Savinov
    \and
    Torsten Sattler
    \and
    Gaurav Chaurasia
}
\institute{
    ETH Zurich
}

\titlerunning{RNN-based Generative Model for Fine-Grained Sketching}
\authorrunning{Jenal \etal}

\maketitle

\begin{abstract}
\input{sections/abstract}
\keywords{
    Generative Adversarial Networks,
    Recurrent Neural Networks,
    Sketching,
    Tree modeling}
\end{abstract}

\input{sections/introduction}
\input{sections/related}
\input{sections/dataset}
\input{sections/methods}
\input{sections/results}

\input{sections/evaluation}
\input{sections/conclusion}

\appendix
\input{sketchgen-supp}

\bibliographystyle{splncs}
\bibliography{bibliography}

\end{document}

%% file: sections/preamble.tex
\usepackage{hyperref}
\usepackage{booktabs}
\usepackage{tabularx}
\usepackage{subcaption}
\usepackage{color}
\usepackage{xspace}
\usepackage{cite}

\makeatletter
\DeclareRobustCommand\onedot{\futurelet\@let@token\@onedot}
\def\@onedot{\ifx\@let@token.\else.\null\fi\xspace}

\def\eg{\emph{e.g}\onedot} 
\def\ie{\emph{i.e}\onedot} 
\def\cf{\emph{c.f}\onedot} 
\def\etc{\emph{etc}\onedot} 
 
\def\etal{\emph{et al}\onedot}
\makeatother

\def\draw{\textsc{draw}\@\xspace}
\def\dcgan{\textsc{dcgan}\@\xspace}
\def\vaegan{\textsc{vae-gan}\@\xspace}

\def\real{\textsc{training}\@\xspace}
\def\fake{\textsc{generated}\@\xspace}
\def\dataset{\textsc{TreeSkel}\@\xspace}
\def\treenet{\textsc{SketchGen}\@\xspace}

\def\x    {x}
\def\xt   {\tilde{x}}
\def\z    {z}
\def\rz   {z_r}
\def\enc  {E}
\def\dec  {G}
\def\dis  {D}
\def\gen  {G}
\def\pxl  {\mathrm{pixel}}
\def\pr   {\mathrm{prior}}
\def\feat {\mathrm{feat}}
\def\pdis {\theta_\dis}
\def\penc {\theta_\enc}
\def\pdec {\theta_\dec}
\def\loss {\mathcal{L}}
\def\ldis {\mathcal{L}_\dis}
\def\lenc {\mathcal{L}_\enc}
\def\lgen {\mathcal{L}_\gen}
\def\lpxl {\mathcal{L}_\pxl}
\def\lpr  {\mathcal{L}_\pr }
\def\lfeat{\mathcal{L}_\feat}
\def\wpxl {\lambda_\pxl}
\def\wfeat{\lambda_\feat}
\newcommand{\brac}[1]{\left(#1\right)}
\newcommand{\sbrac}[1]{\left[#1\right]}
\newcommand{\Enc}[1]{\enc\sbrac{#1}}
\newcommand{\Dec}[1]{\dec\sbrac{#1}}
\newcommand{\Dis}[1]{\dis\sbrac{#1}}
\newcommand{\sigmoid}[1]{\sigma\brac{#1}}
\newcommand{\gauss}[2]{\mathcal{N}\brac{\mathbf{#1},\mathbf{#2}}}
\newcommand{\norm}[1]{\left\lVert#1\right\rVert}
\newcommand{\modulus}[1]{\norm{#1}_2}
\newcommand{\abs}[1]{\left\lvert #1 \right\rvert}
\newcommand{\expect}[1]{\mathbb{E}\sbrac{#1}}

\newcommand{\PAR}[1]{\vskip4pt \noindent{\bf #1~}}

\usepackage{enumitem}

\newenvironment{myitemize}{%
    \begin{itemize} % [noitemsep,topsep=0pt,leftmargin=2em]
        }{%
    \end{itemize}
}

%% file: sections/abstract.tex
Deep generative models have shown great promise when it comes to synthesising novel images. While they can generate images that look convincing on a higher-level, generating fine-grained details is still a challenge. In order to foster research on more powerful generative approaches, this paper proposes a novel task: generative modelling of 2D tree skeletons. Trees are an interesting shape class because they exhibit complexity and variations that are well-suited to measure the ability of a generative model to generated detailed structures. We propose a new dataset for this task and demonstrate that state-of-the-art generative models fail to synthesise realistic images on our benchmark, even though they perform well on current datasets like MNIST digits. Motivated by these results, we propose a novel network architecture based on combining a variational autoencoder using Recurrent Neural Networks and a convolutional discriminator. The network, error metrics and training procedure are adapted to the task of fine-grained sketching. Through quantitative and perceptual experiments, we show that our model outperforms previous work and that our dataset is a valuable benchmark for generative models. We will make our dataset publicly available.

%% file: sections/introduction.tex
\section{Introduction}

\begin{figure}[t]
    \centering
    \begin{subfigure}[c]{0.375\columnwidth}
        \includegraphics[width=\columnwidth]{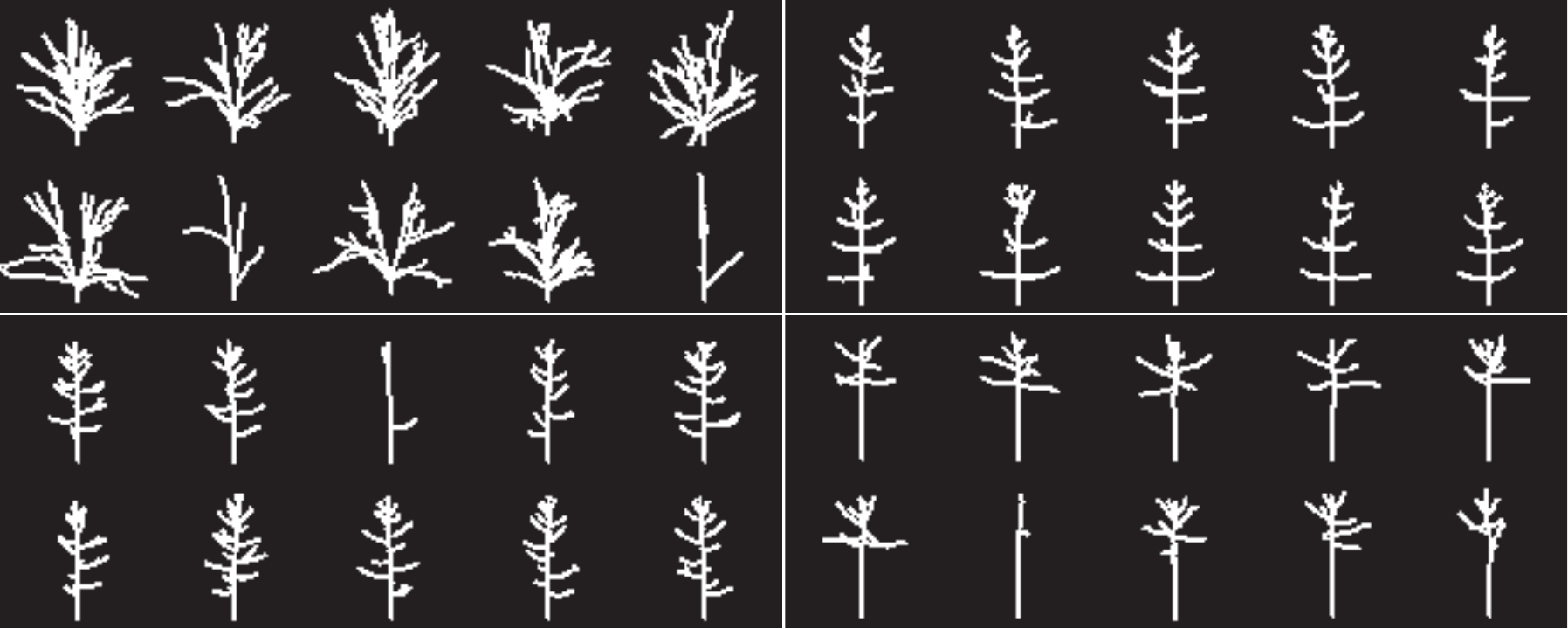}
        \caption{\dataset dataset}
    \end{subfigure}
    \begin{subfigure}[c]{0.3\columnwidth}
        \includegraphics[width=\columnwidth]{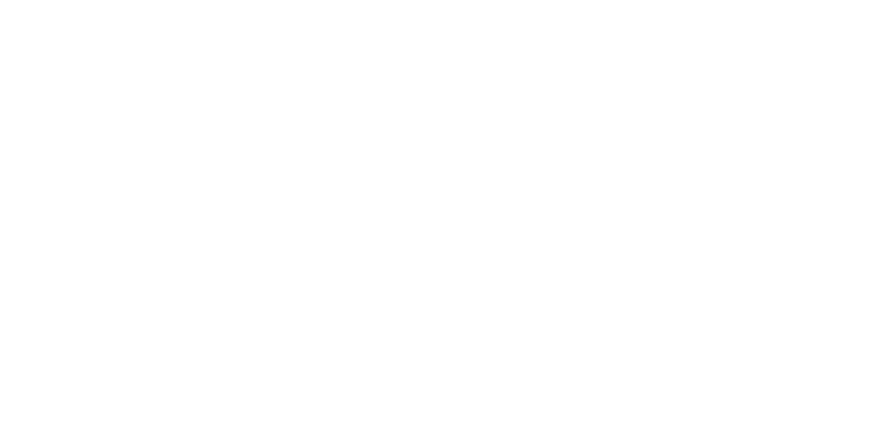}
        \caption{Previous models}
    \end{subfigure}
    \begin{subfigure}[c]{0.3\columnwidth}
        \includegraphics[width=\columnwidth]{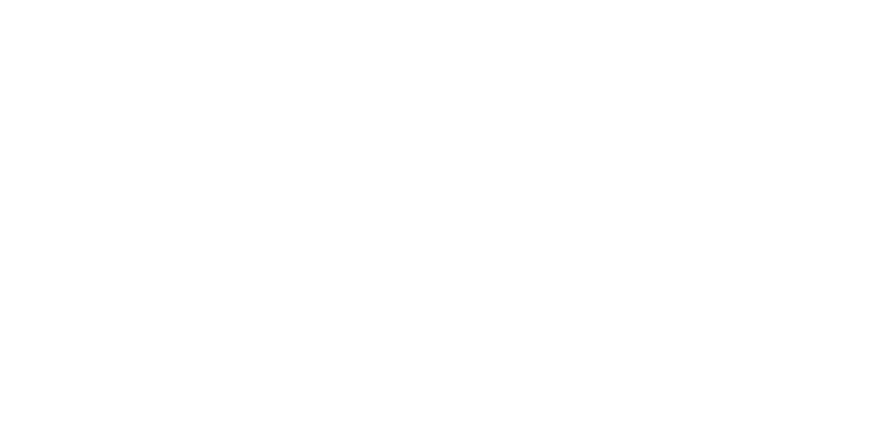}
        \caption{Our results}
    \end{subfigure}
    \caption{Synthesising fine-detailed structures via generative models is an
    interesting problem. We propose new and challenging dataset \dataset in the
    form of 2D trees (a). We also propose a novel generative model \treenet, that produces
    significantly more realistic trees (c) compared to state-of-the-art (b).}
    \label{fig:teaser}
\end{figure}

Generative modelling using deep networks has shown promising results for
synthesising visual content \cite{Dosovitskiy:2014,Radford:2015:dcgan,%
Larsen:2015:vaegan,Dosovitskiy:2016,Wu:2016:3dgan,Bao:2017:cvaegan,%
Nguyen:2017:pnpgan,Odena2017MLR}.
The ability to learn visual semantics makes these models very useful for large
scale synthesis, \eg, for creating assets for graphics applications or for generating training data
for other learning tasks. In this paper, we focus on 2D sketching.
While sketching is  the most basic
content creation task, it is still a challenging problem if fine details need to be produced. The ability to learn how to synthesise fine details from training
sketches is thus a key requirement for generative models for
sketching. % is thus t.

In order to benchmark generative models for the task of fine-grained sketching, we develop a
novel training dataset with complex structures. %  that can be used for benchmarking. %similar
%models, which we will make available.
%
We choose tree skeletons as the shape class for our benchmark (\cf Fig.~\ref{fig:teaser}). % test our network.
 This is in
contrast to the popular choice of man-made objects like tables, chairs \etc in other datasets. Our
motivation is two fold. First, tree modelling is very relevant to high quality
graphics applications and remains a difficult problem in spite of a long line of
research \cite{Lindenmayer:1968,Prusinkiewicz:1994,Weber:1995,Shlyakhter:2001,%
Tan:2007,Xu:2007,Neubert:2007,Tan:2008,Livny:2010,Longay:2012,Bradley:2013,Chaurasia:2017}.
Being able to learn a generative model for trees thus helps to solve a real-world problem. Secondly, trees are an
interesting shape class. They have fine organic details which are hard to
capture for methods tested on structured man-made objects.
We show that state-of-the-art generative models that
produce good results on simple datasets like MNIST hand-drawn digits fail to generate satisfactory results on our dataset
(\cf Fig.~\ref{fig:teaser}).
%produce unsatisfactory results on our dataset.
This highlights the need for
challenging benchmarks for freeform sketching such as the benchmark we propose in this paper. %, which we seek to address.

Motivated by the deficiencies of existing generative models, we propose a new approach able to replicate fine details (\cf Fig.~\ref{fig:teaser}).
We use a Recurrent Neural Network (RNN) to synthesise tree sketches from random points in a low dimensional space, inspired by \draw \cite{Gregor:2015:draw}. The RNN colours
pixels sequentially and connects coloured pixels by varying the size of attention
window to preserve details. We use the RNN as the encoder and decoder of a
variational autoencoder that learns a meaningful low dimensional manifold from
training tree images. This allows generating trees from a particular species as
well as interpolating between species. It also allows reconstructing an exemplar tree sketch.
We train the network in an unsupervised adversarial fashion using a Convolutional
Neural Network (CNN) as discriminator.

%Modelling intra- and
%inter-species variations with hand-crafted features is fairly non-trivial. Deep
%networks are an attractive candidate for learning such variations. In this paper, we therefore propose a learning-based generative model that can replicate fine details.  This task of
%generating trees is well-suited to measure progress in fine-grained 2D sketching,
%while also advancing research in virtual vegetation modelling.
In summary, this paper makes the following contributions:
We propose a novel %Our high-level contributions can be summarised as:
%\begin{myitemize}
%\item a
dataset to benchmark the ability of a generative model to capture fine
    details in 2D sketching together with an evaluation protocol based on improved error metrics. %, for 2D sketching for evaluating sketching quality.
We also propose a network architecture for sketching, together with strategies to alleviate training problems.
%\item a deep network for sketching fine details, and
%\item a protocol for evaluating sketching quality.
%\end{myitemize}
%We motivate the network architecture from the deficiencies of prior models.
We perform extensive quantitatively and perceptual comparisons of our results with
previous models and show that our models is better at recovering fine details. Moreover, our network can learn other classes of shapes, as
demonstrated on MNIST digits.

We will make our new dataset, together with the evaluation protocol, publicly available.

%
%On the technical side, we combine the best ideas in generative modelling: we
%propose improved error metrics for 2D sketching and alleviate training problems.
% The proposed error metrics and training
%procedure are adapted to the task of fine grained 2D sketching. The end result
%is a network capable of scalable unsupervised learning and preserving details.

%% file: sections/related.tex
\section{Related Work}
\label{sec:prevwork}

Generative modelling refers to neural networks that learn the visual or
structural layout of the input and synthesise previously unseen similar samples.
GAN \cite{Goodfellow:2014:gan} is a promising tool for unsupervised learning of
images/shapes in generative models. It uses a \emph{discriminator} to judge the
quality of samples synthesised by the \emph{generator}. Recent work has improved
training \cite{Salimans:2016:gan,Arjovsky:2017:wgan} and evaluation \cite{Theis:2015:ganeval}.

A variety of network architectures have been tested for generative modelling in
various contexts. Dosovitskiy \etal \cite{Dosovitskiy:2014} used CNNs to generate
images of chairs from metadata like view, transformation etc. This was improved
with GANs to generate natural images from random low dimensional points
\cite{Radford:2015:dcgan}.
Larsen \etal \cite{Larsen:2015:vaegan} improve this by adding an autoencoder
\cite{Doersch:2016:autoencoder} which also learns an explicit low dimensional
space. This enables reconstruction \cite{Rezende:2016:3dlearning,Wu:2016:3dgan,Girdhar:2016:object,Li:2017:grass}
and exploration of shape space \cite{Yumer:2015:procedural}. Conditional
generation is described using autoencoders \cite{Sohn:2015:cvae} and GANs
\cite{Nguyen:2017:pnpgan,Bao:2017:cvaegan}. They can learn multiple classes and
generate samples of a particular class by accepting a class label along with
input samples. Latest research on GAN has focused on improving training for
natural image synthesis \cite{Odena2017MLR}.
While all these methods use convolutional networks, Recurrent Neural Networks
(RNN) \cite{Hochreiter:1997:lstm,Gers:2000:rnn} have recently been shown to be
useful for causal pixel synthesis, for digits \cite{Gregor:2015:draw},
natural images \cite{Oord:2016:pxlrnn} and basic 3D shapes \cite{Zou:2017:3drnn}.

Most relevant to our work are recent generative models: \dcgan \cite{Radford:2015:dcgan},
\vaegan \cite{Larsen:2015:vaegan} and \draw \cite{Gregor:2015:draw}. They have
shown impressive results for photorealistic and binary images, and 3D shapes.
This techniques are generic and a good representative for a majority of research
in generative modelling. They are didactic, in that they present a palette of
design choices to compare: CNNs, RNNs, adversarial training, autoencoder \etc,
for designing a new network for a new problem.
Subsequent work has added functionality like conditional generation
\cite{Sohn:2015:cvae,Bao:2017:cvaegan,Nguyen:2017:pnpgan}; our dataset has only
one class of shapes, so we do not require conditional generation.
We show the results of \dcgan, \vaegan and \draw on
our dataset (\cf Sec. \ref{sec:dataset:prevwork}) and design our network after
analysing the their artefacts (\cf Sec.~\ref{sec:treenet}). We later compare
our results to these using quantitative evaluation and perceptual studies
(\cf Sec.~\ref{sec:eval}).

%% file: sections/dataset.tex
\section{\dataset: A Large Dataset for Generative Modelling of Fine Details}
\label{sec:dataset}

\begin{figure}[t]
    \centering
    \begin{subfigure}[c]{0.2\columnwidth}
        \includegraphics[width=\columnwidth]{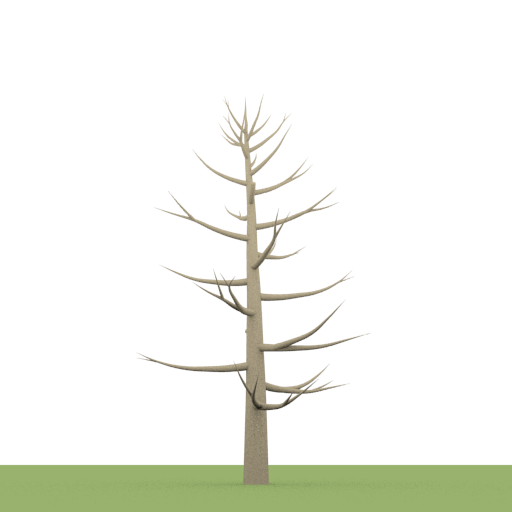}
    \end{subfigure}
    \begin{subfigure}[c]{0.2\columnwidth}
        \includegraphics[width=\columnwidth]{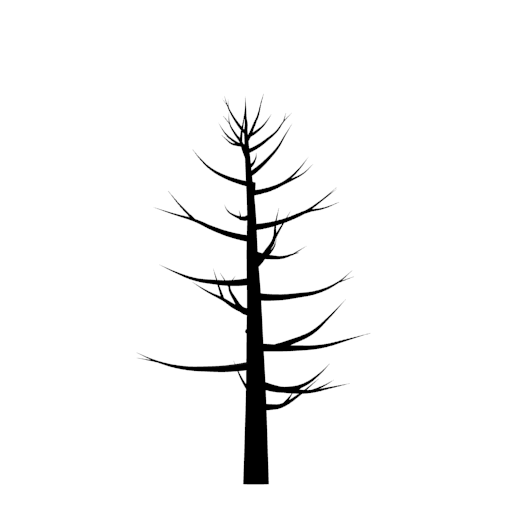}
    \end{subfigure}
    \begin{subfigure}[c]{0.2\columnwidth}
        \includegraphics[width=\columnwidth]{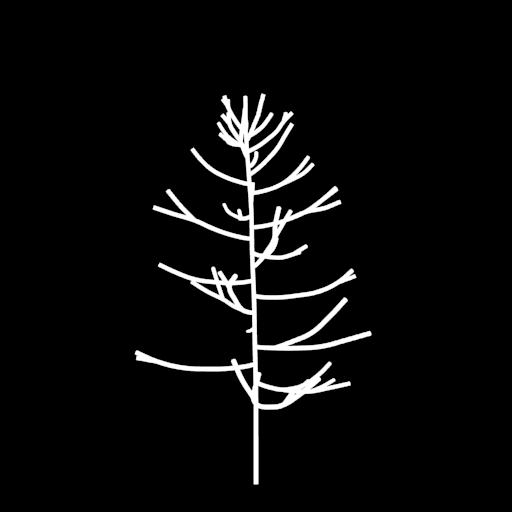}
    \end{subfigure}
    \caption{\dataset creation. From left to right: 3D Larch tree model created
        in Blender, binary rendering of tree, and tree skeleton after median
        axis computation. \dataset contains 15000 tree skeletons of 15 species.}
    \label{fig:dataset}
\end{figure}

As one of the main contributions of this paper, we create \dataset, a new dataset consisting of tree skeletons.
In the following, we motivate our new dataset, describe its creation, and demonstrate that it is challenging for state-of-the-art generative models. % towards the goal of sketching.
We will make this dataset publicly available upon publication.

We decided to create a dataset of trees as they
%Trees
have interesting intra-class variations and exhibit fine details.
In addition, a tree-based dataset also address the need for
virtual vegetation in graphics applications. \dataset consists of 15000 sketches, with
1000 sketches for each of its 15 different tree species. The sketches were created in
an artist assisted process described below. \dataset is designed as a response to the lack of %to fill a gap in the set of existing
benchmark datasets for fine grained sketching. Existing datasets
in deep learning are geared towards other tasks. % specific targets. % and do not address this
%task.
For example, CIFAR is designed for object classification \cite{Krizhevsky:2009:learning},
ImageNet for image classification \cite{ILSVRC15}, and MNIST for handwriting
recognition \cite{MNIST15}. %, \etc.
ImageNet and CIFAR model the distribution of natural images, and MNIST is a
collection of natural variations in handwritten digits.
While MNIST is a valid training set for sketching, the structure of handwritten digits is rather simplistic. As such, MNIST is not a suitable benchmark to evaluate the ability of generative models to produce fine details. % too simple to teach complex structure.
In contrast, \dataset is
deliberately curated to teach a network how to sketch fine details and
variations. As such, it is similar in spirit to Gharbi \etal \cite{Gharbi:2016} who
proposed a dataset to alleviate specific image processing artefacts.
Our dataset is the first that can be used for
sketching because:
\begin{myitemize}
\item it abstracts away texture, lighting, \etc and focuses on the skeletal
    structures most relevant for sketching,
\item it is created under controlled conditions to exhibit fine details and a
    large gamut of variations so as to test a generative model's expressive ability,
\item it contains equal number of samples of each species so as to make the
    network equally proficient at all species,
\item it resembles real trees so as to retain relevance to applications that
    require virtual vegetation.
\end{myitemize}

In order to create the dataset, we selected 15 tree species as the classes in \dataset: Acacia, Beech, Callistemon, Cedar, Chestnut,
Elm, Japanese Maple, Kauri, Larch, Linden, Pine, Quaking Aspen, Small Maple,
Teak and White Birch. These species are sufficiently diverse to test the
expressive ability of the network. We asked artists to design one or more 3D
tree models for each species using the \textsc{Sapling}\footnote{\small{\url{https://github.com/abpy/improved-sapling-tree-generator}}}
addon in Blender\footnote{\small{\url{https://www.blender.org/}}}. This is a
time consuming process because modelling tools have a number to parameters to
tune for a specific appearance. Given the models, we thus added controlled randomisation to each
parameter in the modelling tool to automatically generate a large number of tree models. We
then render the models %to 2D sketches
from four randomly chosen viewpoints into
64$\times$64 binary images (\cf Fig.~\ref{fig:dataset}). The viewpoints are
placed at average human height with a field of view that captures the entire
tree. We extract the medial axis which represents the tree skeleton. This allows
the network to focus on structure; branch thickness can be conveniently handled
in post-process if needed. We also restrict the the number of times the trunk
splits into branches from the root to the tip of a twig to 2 while creating 3D
trees. Higher splitting factors require rendering to image resolutions that are
currently computationally intractable for deep networks. These final images
constitute the database (\cf Fig.~\ref{fig:teaser}). In the following, we refer to samples from \dataset as
\real and network synthesised samples as \fake.

\subsection{Challenges Contained in The Dataset}
\label{sec:dataset:prevwork}

\begin{figure}[t]
    \begin{subfigure}[t]{0.323\textwidth}
        \includegraphics[width=\columnwidth]{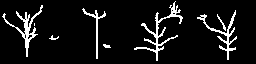}
        \caption{\dcgan: disconnected \& unnatural branches}
        \label{fig:dcgan}
    \end{subfigure}
    \hfill
    \begin{subfigure}[t]{0.323\textwidth}
        \includegraphics[width=\columnwidth]{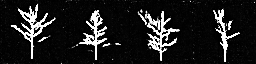}
        \caption{\vaegan: disconnected branches \& high-frequency noise}
        \label{fig:vaegan}
    \end{subfigure}
    \hfill
    \begin{subfigure}[t]{0.323\textwidth}
        \includegraphics[width=\columnwidth]{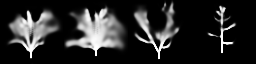}
        \caption{\draw: blurred results}
        \label{fig:draw}
    \end{subfigure}
    \caption{Evaluation of state-of-the-art generative models on our novel \dataset dataset. As can be seen, \dcgan~\cite{Radford:2015:dcgan}, \vaegan~\cite{Larsen:2015:vaegan}, and \draw~\cite{Gregor:2015:draw} each have distinct failure modes and none of them generates realistic tree sketches. }
    \label{fig:prevresults}
\end{figure}

In order to demonstrate the challenges inherent to fine-grained sketching of trees that are covered by our dataset, we tested state-of-the-art generative models (\cf Sec.~\ref{sec:prevwork}) on \dataset. % and observed prominent artifacts.
Exemplary results from this experiment are shown in Fig.~\ref{fig:prevresults}:
\dcgan~\cite{Radford:2015:dcgan} produces unnaturally looking structures and
disconnected branches (\cf Fig.~\ref{fig:dcgan}). \vaegan~\cite{Larsen:2015:vaegan}
results are better but they have far more disconnected branches, to the point
that they resemble high frequency noise (\cf Fig.~\ref{fig:vaegan}). These results
show that our dataset contains sufficient details that cannot be modelled by a
convolutional network that does not synthesise pixels in a causal order.
Random
structures can pop up and the network has no mechanism to ensure creating a connectivity
that resembles valid images. \draw~\cite{Gregor:2015:draw} uses a RNN
encoder-decoder to learn a pixel colouring order during sketching. This results in less
noise and fewer disconnected branches, but heavily blurs fine details
(\cf Fig.~\ref{fig:draw}). All these artefacts: unnatural structures,
disconnected structures, noise, blur \etc, are not observed when evaluating on existing datasets such as % cannot be seen with current datasets
%like
 MNIST digits. \dataset is thus a challenging dataset that exposes the
limitations of existing generative models.

%% file: sections/methods.tex
\section{\treenet: A Generative Model}
\label{sec:treenet}

\begin{figure}[ht]
    \centering
    \includegraphics[width=\columnwidth]{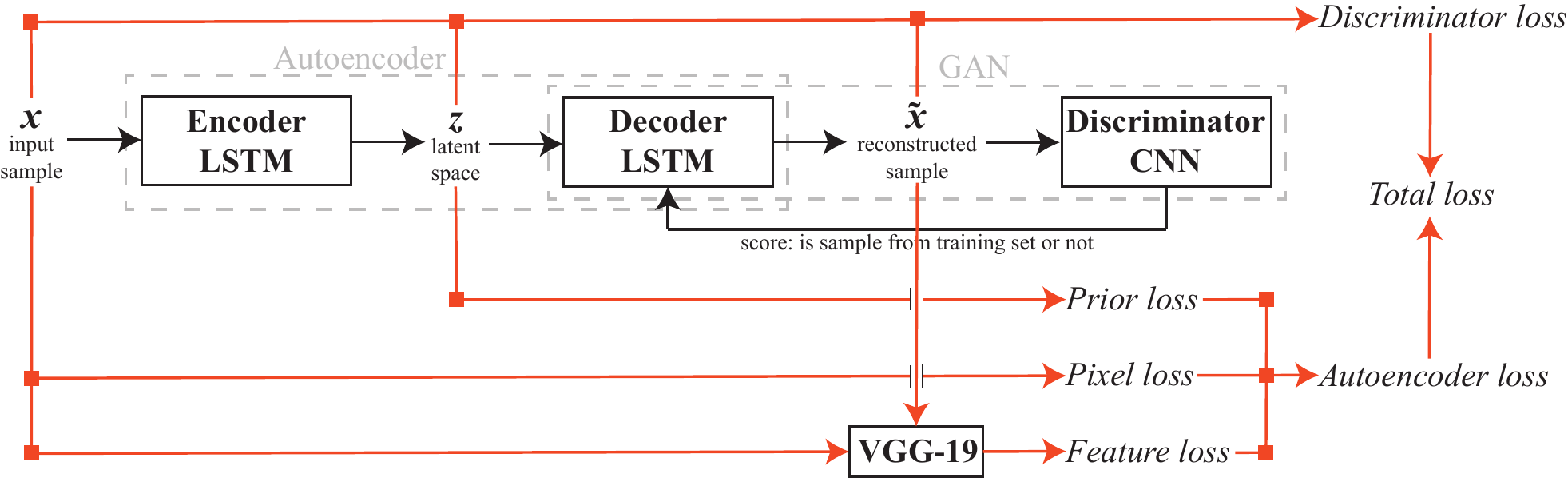}
    \caption{\treenet architecture with loss functions.}
    \label{fig:treenet}
\end{figure}
\begin{figure}[ht]
    \centering
    \includegraphics[width=\columnwidth]{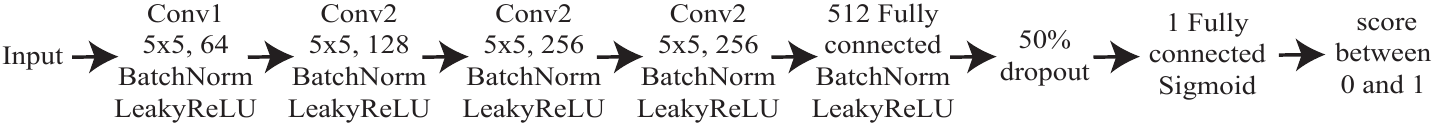}
    \caption{\treenet discriminator CNN layers.}
    \label{fig:discriminator}
\end{figure}

\begin{figure}[t]
    \centering
    \begin{subfigure}{0.3\textwidth}
        \includegraphics[width=\columnwidth]{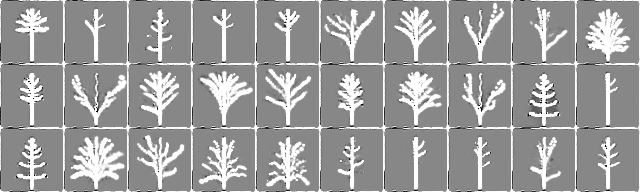}
        \caption{Only $\lfeat$}
    \end{subfigure}
    \begin{subfigure}{0.3\textwidth}
        \includegraphics[width=\columnwidth]{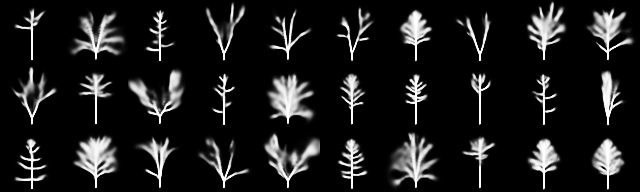}
        \caption{Only $\lpxl$}
    \end{subfigure}
    \begin{subfigure}{0.3\textwidth}
        \includegraphics[width=\columnwidth]{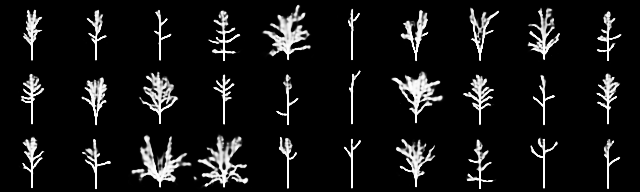}
        \caption{$\lfeat$ and $\lpxl$}
    \end{subfigure}
    \caption{Ablation study: results using only feature loss, only pixel loss, and
    both losses. Only the feature loss can give noisy results and arbitrary pixel
    values. Using both losses preserves features and gives sharper results than
    pixel loss alone.}
    \label{fig:featureloss}
\end{figure}

The poor results of previous generative models (\cf Fig.~\ref{fig:prevresults})
demonstrate the need for a better model. We propose the \treenet model that is
better adapted to fine-grained sketching. Our network consists of three parts.
The first is an \emph{encoder} that converts input image into a point in a latent space. The second is a \emph{decoder} that converts a point in latent space back into an image. The last is a \emph{discriminator} that classifies an image as one
belonging to the training dataset or not. The encoder-decoder combine as an
\emph{autoencoder} that enables reconstruction, and the decoder-discriminator
combine as a GAN that allows unsupervised learning. Our architecture is illustrated in Fig.~\ref{fig:treenet}.

An important difference between our network and previous models is that we use a
Recurrent Neural Network (RNN) as the encoder and decoder, in contrast to the CNNs
used by previous models \cite{Dosovitskiy:2014,Radford:2015:dcgan,Larsen:2015:vaegan,%
Dosovitskiy:2016,Bao:2017:cvaegan,Nguyen:2017:pnpgan,Odena2017MLR}
This choice is motivated by the poor sketching results of convolutional decoders
on \dataset (\cf Fig.~\ref{fig:prevresults}). We attribute this failure to
the lack of pixel colouring order in convolutional networks, which leaves
disconnected structures and high frequency noise. An RNN is a promising solution
that can learn causal pixel synthesis. Incorporating an autoencoder is consistent
with recent generative models because it creates a meaningful latent space that
can be used for reconstruction from exemplars.
In addition, it enables interpolation in the latent space for controllable synthesis \cite{Rezende:2016:3dlearning,Wu:2016:3dgan,%
Girdhar:2016:object,Li:2017:grass,Yumer:2015:procedural}.

In order to ensure that our synthesised samples contain fine-grained details while still being semantically consistent, \eg, to avoid disconnected branches, we measure both pixel-level difference as well as higher-level feature similarity in our loss functions. % to compare
%semantic similarity between images and their reconstructions. Pixelwise difference
%alone is not sufficient for measuring structural similarity of two sketches.

\subsection{Network details}

Our \textbf{encoder} is a LSTM \cite{Hochreiter:1997:lstm} that takes the input
sample and the hidden vector output of the previous time step to produce the output
of the current time step. The details of the architecture are the same as those
in \draw \cite{Gregor:2015:draw}, with different hyperparameters
(\cf Table~\ref{tab:hyperparams}). We also incorporate the attention mechanism
from \draw to determine the window of pixels to be sketched at any
time step. Our experiments showed better results with
the attention mechanism. The \textbf{decoder} mirrors the same LSTM architecture.

Our \textbf{discriminator} is a multi-layer CNN (\cf Fig.~\ref{fig:discriminator}).
All convolutions are 5$\times$5 and followed by batch normalisation
\cite{Ioffe:2015:batchnorm} and leaky ReLU \cite{Maas:2013:leakyrelu}.
The convolutional layers are followed by fully connected layers, a 50\% dropout
layer to prevent overfitting, and a sigmoid that returns the likelihood that the
input image belongs to the training dataset. We did not use a very deep network
like VGG-19 \cite{Simonyan:2015:vgg} because our discriminator only has to decide
whether binary sketches are from the training dataset or not, which is simpler
than VGG's task of learning natural images.

% ------------------------------------------------------------------------------

\subsection{Loss Functions}

We train the autoencoder and the GAN simultaneously. The overall objective
is a combination of the autoencoder loss $\lenc$ and discriminator loss $\ldis$:
\begin{equation}
    \loss = \lenc + \ldis \enspace .
    \label{eq:loss}
\end{equation}
The autoencoder loss $\lenc$ ensures that the encoder and decoder are trained
such that input images to the encoder and their corresponding decoded outputs
retain fidelity. The discriminator loss $\ldis$ additionally aims to ensure that the output
of the decoder is always indistinguishable from training set samples,
irrespective of the input to the decoder. Without $\ldis$, the only latent space
points that can be decoded to images that look like training set are those that
are close to encoding of training set images while random points in latent space  may
decode to arbitrary results. $\ldis$ is therefore indispensable for synthesising
sketches from random latent space points. We now detail the autoencoder loss while the
discriminator loss is described at the end of the section.

We formulate the autoencoder loss $\lenc$ as a combination of three terms, which
ensure input-output fidelity and enforce a meaningful latent space:
\begin{equation}
    \lenc = \lpr + \wpxl \lpxl + \wfeat \lfeat \enspace .
    \label{eq:encloss}
\end{equation}
Here, $\lpxl$ and $\lfeat$ are pixel and feature losses which directly measure
the fidelity between input image to the encoder and output image of the decoder.
$\lpr$ is the prior loss which regularises $\lpxl$ and $\lfeat$ such that a
meaningful latent space is constructed. Without $\lpr$, the latent space may
assume an arbitrary distribution and the purpose of having separate encoder and
decoder will be lost.

In the following, we denote the input training samples as  $\x$, and the outputs of the encoder, decoder, and discriminator as
$\Enc{\cdot}$, $\Dec{\cdot}$, and $\Dis{\cdot}$.
Using these operators, we can express the latent space conversion of the input
image as $z=\Enc{\x}$, and the reconstruction result as $\xt=\Dec{\Enc{\x}}$. We
denote a randomly sampled point in latent space as $\rz$. We now explain each loss
function in detail.

The \textbf{prior loss}
$\lpr$ regularises the encoder by enforcing a probability distribution known
\textit{a priori} on the latent space. We enforce a zero mean Gaussian
distribution. As formulated by Kingma \& Welling \cite{Kingma:2014:vae}, we compute
the analytic KL-divergence between the $k$-dimensional latent space
$p\brac{\z|\x}$ and the prior $\gauss{0}{I}$ using the mean $\mu$ and covariance
matrix $\Sigma$ of the observed latent space:
\begin{equation}
    \lpr = \frac{1}{2} \brac{\mathrm{tr}\brac{\Sigma} + \mu^T\mu - k - \log\brac{\det\brac{\Sigma}}} \enspace .
    \label{eq:priorloss}
\end{equation}

The \textbf{pixel loss}
$\lpxl$ is a direct measure of fidelity between input $x$ and reconstruction $\xt$.
To this end, we use pixel-wise mean squared error.
\begin{equation}
    \lpxl = \modulus{\x-\xt} \enspace .
    \label{eq:pixelloss}
\end{equation}

The pixel loss $\lpxl$ only accounts for low-level similarity. In order to synthesise realistic tree shapes, it is necessary to also account for higher-level similarity.
We thus also use a \textbf{high-level feature loss} that accounts for consistent branching behaviour and connectivity.
%We seek to measure semantic similarity of two sketches, for which pixelwise
%difference is not an appropriate metric.
Intermediate representations learnt by
classifier CNNs have been shown to encode high-level features that can be
used to measure similarity between images \cite{Aittala:2016:texsyn,Gatys:2015:style}.
Generative models have also used features extracted from intermediate layers of
pre-trained CNNs \cite{Dosovitskiy:2016,Larsen:2015:vaegan}, or discriminator
\cite{Bao:2017:cvaegan}.
We extract feature maps as the intermediate activations of a pre-trained VGG-19 network~\cite{Simonyan:2015:vgg}. Let $V_i$ and
$\tilde{V}_i$ denote the activations of the $i^\mathrm{th}$ convolutional plus ReLU
layer of VGG applied to input $\x$ and reconstruction $\xt$, respectively. The
feature loss $\lfeat$ is given by:
\begin{equation}
    \lfeat = \frac{1}{N} \sum_{i\in L} \sigmoid{\frac{i}{\abs{L}}} \modulus{V_i-\tilde{V}_i} \enspace .
    \label{eq:featloss}
\end{equation}
Here $\sigmoid{\cdot}$ is the sigmoid function,
$N = \sum_{i} \sigmoid{\frac{i}{\abs{L}}}$ is a normalisation factor, and
$L$ is the set of VGG layers whose activation is used.

In order to account for various kinds of features, we use multiple early VGG
layers $L = \left\{\text{ReLU2\_2, ReLU3\_4, ReLU4\_4, ReLU5\_4} \right\}$. The
progressively increasing sigmoidal weights prioritise semi-global features. This
loss preserves structure and alleviates blurring compared to the pixel loss
(\cf Fig.~\ref{fig:featureloss}). However, it alone is not sufficient because
it can lead to high frequency artefacts \cite{Mahendran:2016:cnnvis}. We combine
it with the pixel loss using weights $\wfeat=10$ and $\wpxl=0.5$ in Eq.~\ref{eq:encloss}.

Our \textbf{discriminator loss}
is formulated as Wasserstein GAN objective \cite{Arjovsky:2017:wgan}, which reportedly is more stable than the cross-entropy GAN objective used by
\vaegan:
\begin{equation}
    \ldis = \expect{-\Dis{\x} + \Dis{\Dec{\rz}} + \Dis{\xt}} \enspace .
    \label{eq:ganloss}
\end{equation}
Here, $\expect{\cdot}$ denotes the expectation over all training samples $\x$ in a
mini-batch, their corresponding reconstructions $\xt$, and an equal number of
randomly drawn points $\rz$.

We also include the discriminator result for the decoded
training samples $\Dis{\xt}$ in Eq.~\ref{eq:ganloss}, in
addition to the usual terms involving $\x$ and $\rz$. The intuition is as
follows: The traditional $\Dis{\x}$ and $\Dis{\Dec{\rz}}$ terms measures the
ability of the discriminator to judge if the image decoded from a random latent
point $\Dec{\rz}$ is similar to a training image $\x$. By adding $\Dis{\xt}$, we
are judging the quality of the discriminator by its ability to distinguish not
only random synthesised samples from training images, but also reconstructions
of training samples from training samples themselves. Using this addition improved our results.
% $\ldis$ varies from $-1$ for a strong discriminator and $0.5$ for a weak one.

% ------------------------------------------------------------------------------

\subsection{Training Details}

Training GANs is non-trivial because a discriminator that becomes too powerful
too early prevents a proper training of the generator. We alleviate this by
selectively training the discriminator so that it never gets too strong compared
to the generator \cite{Larsen:2015:vaegan,Dosovitskiy:2016}. We define a \textbf{generator
loss} $\lgen = -\expect{\Dis{\Dec{\rz}} + \Dis{\xt}}$ as the opposite of $\ldis$
(Eq.~\ref{eq:ganloss}). We omit the $\x$ term because $\lgen$ only measures
the likelihood of the discriminator being fooled by the decoder.
% It varies from $-1$ for a strong generator to $0$ for a weak generator.
In practice, $\frac{\ldis}{\lgen}$ should be close to $1$, indicating that the
generator is not much stronger than the discriminator. We thus skip a discriminator
update if
\begin{equation}
    \abs{{\ldis}/{\lgen}} \ge 1.5 \log\brac{1+j} \enspace,
    \label{eq:ganupdate}
\end{equation}
where $j$ is the number of the current epoch.
By choosing the threshold as an increasing function of the training epoch, we
ensure that the discriminator parameters are updated rarely in the beginning
and frequently afterwards when the generator has received some training. Encoder
and decoder parameters are always updated.

In practice, we do not update all network parameters w.r.t. the combined loss
(Eq.~\ref{eq:loss}) during optimisation. We update the parameters of the encoder
$\penc$ by backpropagating the gradient of the autoencoder loss, the decoder $\pdec$ by the
autoencoder and generator loss, and the discriminator $\pdis$ by its own loss:
\begin{equation}
    \penc \xleftarrow{+} \nabla_{\penc}\brac{\lenc}  \enspace ,      \quad
    \pdec \xleftarrow{+} \nabla_{\pdec}\brac{\lgen+\lenc} \enspace , \quad
    \pdis \xleftarrow{+} \nabla_{\pdis}\brac{\ldis} \enspace .       \label{eq:update}
\end{equation}
This implementation strategy was also used by Larsen \etal \cite{Larsen:2015:vaegan}.
We use ADAM \cite{Kingma:2015:adam} for optimisation with momentum $\beta_1=0.5$.
We also clip the gradient norm at $10$ in all losses to avoid exploding gradients
because of accumulation over long time sequences in LSTMs \cite{pascanu2013difficulty}.

\begin{table}
    \caption{Hyperparameters of our approach}
    \label{tab:hyperparams}
    \centering
    \begin{footnotesize}
%    \footnotesize{
    \begin{tabular}{@{}p{0.38\columnwidth}@{}p{0.08\columnwidth}|p{0.38\columnwidth}@{}p{0.09\columnwidth}}
        \toprule
        \textbf{Training}         &           & \textbf{Encoder-decoder} &                   \\
        epochs                    & $600$     & learning rate            & $5$$\times$$10^{-4}$ \\
        mini-batch size           & $100$     & read attention window    & $7$$\times$$7$    \\
        \textbf{Discriminator}    &           & write attention window   & $12$$\times$$12$  \\
        learning rate             & $10^{-4}$ & LSTM time steps          & $64$              \\
        leaky ReLU slope          & $0.2$     & LSTM hidden vector size  & $256$             \\
        batch norm decay          & $0.9$     & latent space dimensionality  & $100$         \\
        \bottomrule
    \end{tabular}
    \end{footnotesize}
%    }
\end{table}

%% file: sections/results.tex
\section{Evaluation}
In the following, we evaluate \treenet and compare its performance against state-of-the-art methods on our \dataset dataset.
Sec.~\ref{sec:results} presents a qualitative evaluation of \treenet.
Sec.~\ref{sec:eval} then provides quantitative results using both automated metrics as well as a user study.
%We present qualitative results of \treenet and then quantitative comparisons
%with previous generative models on sketching task using our \dataset dataset.

\subsection{Qualitative Evaluation}
\label{sec:results}

\begin{figure}[t]
    \centering
    \begin{subfigure}{\textwidth}
        \includegraphics[width=\columnwidth]{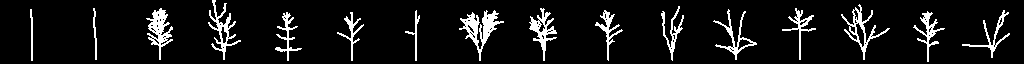}
    \end{subfigure}
    \begin{subfigure}{\textwidth}
        \includegraphics[width=\columnwidth]{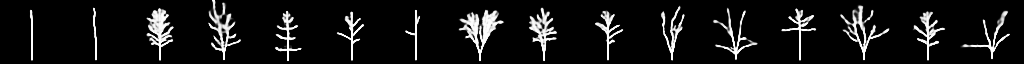}
    \end{subfigure}
    \caption{Reconstruction. Even though \treenet has never seen the top row
    samples during training, it can faithfully reconstruct them (bottom).}
    \label{fig:reconstruct}
\end{figure}

\begin{figure}[t]
    \centering
    \includegraphics[width=0.7\textwidth]{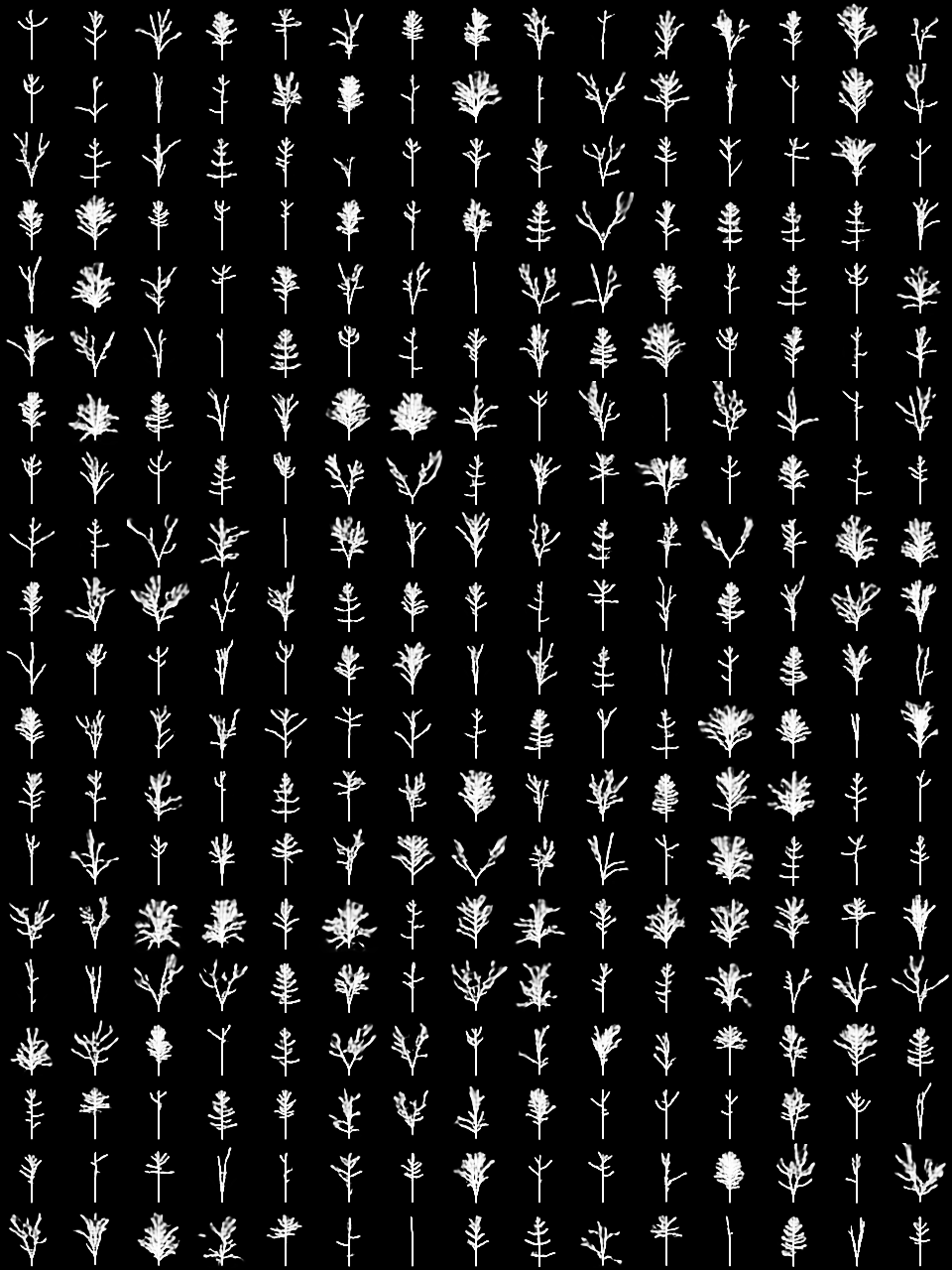}
    \caption{Samples synthesised by the decoder from random latent space points.
    Details best viewed at 500\% zoom.}
    \label{fig:results}
\end{figure}

\begin{figure}[t]
    \centering
    \begin{subfigure}[t]{0.47\textwidth}
        \includegraphics[width=\columnwidth]{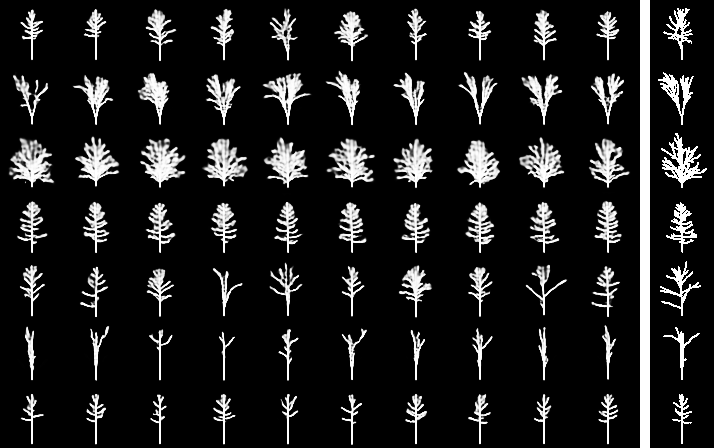}
        \caption{\treenet does not copy but learns: training samples (right) differ significantly from their 10 most similar randomly generated samples (left).}
        \label{fig:closestmatch}
    \end{subfigure}
    \hfill
    \begin{subfigure}[t]{0.47\textwidth}
        \includegraphics[width=\columnwidth]{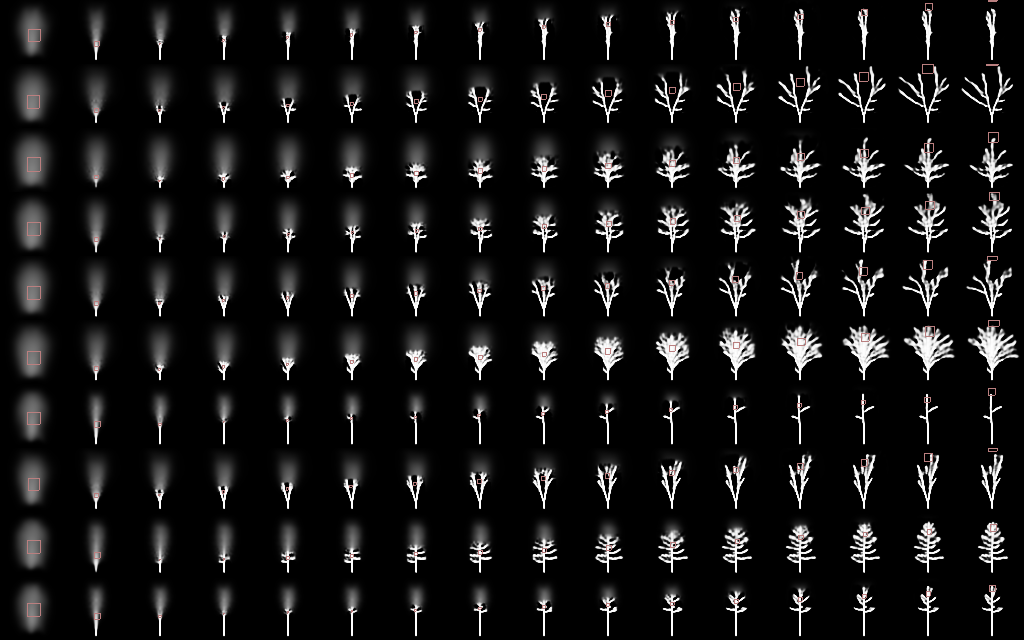}
        \caption{Temporal sketching by the RNN in \treenet.}
        \label{fig:sketchsequence}
    \end{subfigure}
    \caption{\treenet results. Best viewed at 500\% zoom.}
\end{figure}

% \begin{figure}
%     \begin{subfigure}{0.09\columnwidth}
%         \centering
%         \includegraphics[width=\columnwidth]{figures/treenet/reconstruct_species_species_templates.png}
%     \end{subfigure}
%     \hfill
%     \begin{subfigure}{0.9\columnwidth}
%         \centering
%         \includegraphics[width=\columnwidth]{figures/treenet/reconstruct_species_reconstructions.png}
%     \end{subfigure}
%     \caption{Species specific synthesis. Training sample of a species (left) and
%         synthesised samples (right) by adding small perturbation to latent space
%         encoding of the training sample. The results appear to be same species
%         which shows that species are clustered in latent space.}
%     \label{fig:speciesreconstruct}
% \end{figure}

In our qualitative evaluation of \treenet, we are interested both in a comparison with existing methods as well as a deeper understanding of \treenet's abilities, especially its ability to synthesise novel tree samples.

First, we demonstrate the ability of \treenet's decoder to reconstruct trees not contained in the training sample.
This is shown in Fig.~\ref{fig:reconstruct}, where we first use \treenet's encoder to map a previously unseen sample to a point in the latent space and then use the decoder to reconstruct the tree starting from this point. %The network can reconstruct an exemplar tree skeleton at high fidelity: we
%first encode the exemplar to a latent space point, and then decode to a sketch
%(\cf Fig.~\ref{fig:reconstruct}).
As can be seen, \treenet  faithfully reconstructs these examples.

Fig.~\ref{fig:results} shows trees synthesised by \treenet's decoder from randomly selected points in the latent space.
As can be seen, \treenet is able to generate
%We present qualitative analysis of the network: we show synthesised samples,
%reconstructions and and latent space analysis. \treenet decodes random latent
%space points to
tree images with detailed branching structures.
%(\cf Fig.~\ref{fig:results}).
Visual inspection shows that these results do not
have the implausible branch structures of \dcgan, high-frequency noise of
of \vaegan and blurry structures of \draw (compare Fig.~\ref{fig:prevresults}
with Fig.~\ref{fig:results}).
These results clearly show that \treenet generates more realistically looking trees.

The samples synthesised by \treenet cover all 15
species,  showing that the network is not biased towards any particular shape.
However, it remains to be shown that its results are not just simple copies of training samples, but that \treenet actually learns about plausible tree structures.
We demonstrate this through two experiments:
First, we generated around 1000 synthetic trees, picked random training samples, and for each sample retained its
10 pixel-wise closest matches among the synthesised trees.
Fig.~\ref{fig:closestmatch} compares training samples with these most similar synthesised trees. As can be seen, each synthesised tree differs from the training samples in details.

\begin{figure}[h]
    \centering
    \begin{subfigure}{0.013\textwidth}
        \centering
        \rotatebox{90}{\tiny Pine to}
    \end{subfigure}
    \begin{subfigure}{0.013\textwidth}
        \centering
        \rotatebox{90}{\tiny Chestnut}
    \end{subfigure}
    \begin{subfigure}{0.45\textwidth}
        \centering
        \includegraphics[width=\columnwidth]{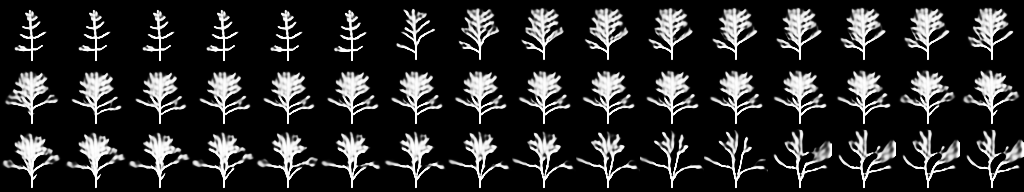}
    \end{subfigure}
    \hfill
    \begin{subfigure}{0.013\textwidth}
        \centering
        \rotatebox{90}{\tiny Beech to}
    \end{subfigure}
    \begin{subfigure}{0.013\textwidth}
        \centering
        \rotatebox{90}{\tiny Acacia}
    \end{subfigure}
    \begin{subfigure}{0.45\textwidth}
        \centering
        \includegraphics[width=\columnwidth]{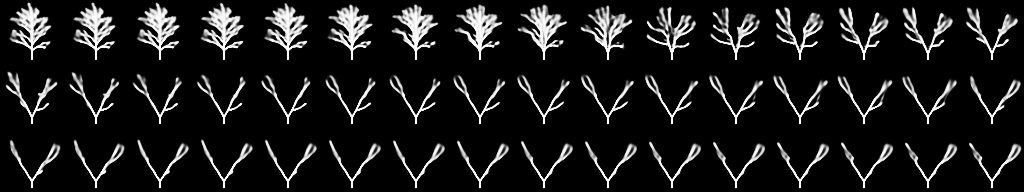}
    \end{subfigure}
    \\
    \begin{subfigure}{0.013\textwidth}
        \centering
        \rotatebox{90}{\tiny Kauri to}
    \end{subfigure}
    \begin{subfigure}{0.013\textwidth}
        \centering
        \rotatebox{90}{\tiny Callistemon}
    \end{subfigure}
    \begin{subfigure}{0.45\textwidth}
        \centering
        \includegraphics[width=\columnwidth]{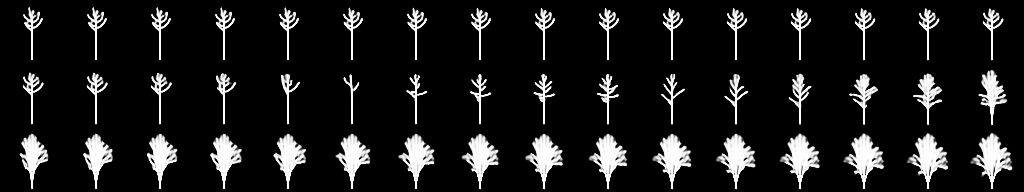}
    \end{subfigure}
    \hfill
    \begin{subfigure}{0.013\textwidth}
        \centering
        \rotatebox{90}{\tiny Linden to}
    \end{subfigure}
    \begin{subfigure}{0.013\textwidth}
        \centering
        \rotatebox{90}{\tiny Acacia}
    \end{subfigure}
    \begin{subfigure}{0.45\textwidth}
        \centering
        \includegraphics[width=\columnwidth]{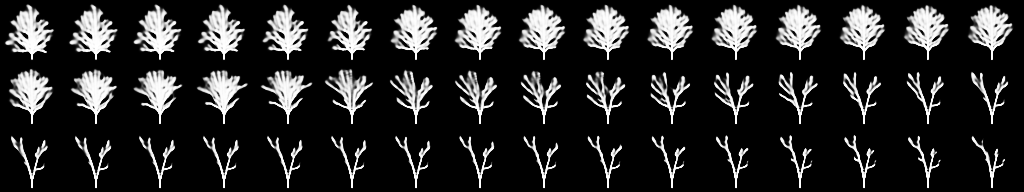}
    \end{subfigure}
    \caption{Latent space interpolation. We choose two random tree samples of
        different species, map them to the latent space, and show the decoded
        results from interpolated latent space points. These show that \treenet
        learns a meaningful latent space. Best viewed at 500\% zoom.}
    \label{fig:latentinterp}
\end{figure}

% results, none of which is a copy of the
%training sample (\cf Fig.~\ref{fig:closestmatch}).
% We use a similar process for generating samples
% of a particular species: we pick an exemplar of that species, encode it to
% latent space point, add small noise to the point and decode it to a new sketch.
% New samples appear to be the same species, but never copies of the exemplar
% (\cf Fig.~\ref{fig:speciesreconstruct}).
% This suggests that different species are spatially clustered in the latent space.
As a second experiment, we linearly interpolate between points in the latent space.
The endpoints are obtained by passing images of two different species
through the encoder.
As shown in Fig.~\ref{fig:latentinterp}, \treenet decodes the
%The latent space learnt by the network is meaningful: we linearly interpolate
%between latent space points obtained by passing images of two different species
%through the encoder (\cf Fig.~\ref{fig:latentinterp}). I
interpolated points to plausible tree shapes, showing a gradual change from one species to another. This clearly demonstrates \treenet's ability to automatically learn meaningful relations between the different species.

\begin{figure}[t]
    \centering
    \begin{subfigure}{0.35\columnwidth}
        \centering
        \includegraphics[width=\columnwidth]{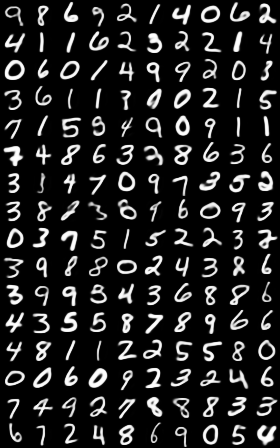}
    \end{subfigure}
    \begin{subfigure}{0.35\columnwidth}
        \centering
        \includegraphics[width=\columnwidth]{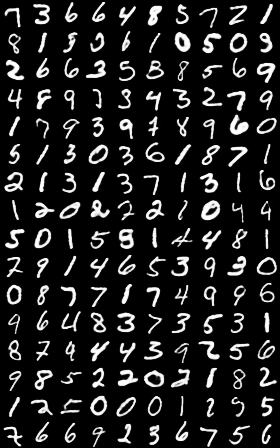}
    \end{subfigure}
    \caption{\treenet generalises beyond trees. Its results (right) on MNIST
    digits are on par with the original results in \draw (left).}
    \label{fig:mnist}
\end{figure}

While the latent space captures high-level structure, fine details are handled
by the attention mechanism of the RNN inside the autoencoder. Visualising
the sketching sequence over time shows the temporal dependency and evolution of
the attention mechanism (\cf Fig.~\ref{fig:sketchsequence}). \treenet learns
this in an unsupervised manner.
We also tested \treenet on MNIST digits \cite{MNIST15} and found the results
on par with prior models (\cf Fig.~\ref{fig:mnist}). This shows that the network architecture of \treenet generalises to other datasets.
%is not dataset specific.

Finally, we perform an ablation study to understand the impact of different parts of our loss function. We obtain the best results using a weighted combination of pixel and feature loss. Using only the feature loss, there is no mechanism to constrain actual pixel values. Thus, a linear or some other transformation of the pixel intensities can give a low feature loss. Using only the pixel loss results in more blur than feature and pixel loss combined. This is because under the $\ell_2$ norm, blurry samples still give low pixel losses and there is no mechanism to enforce sharp results. These results are demonstrated in Fig.~\ref{fig:featureloss}.

%% file: sections/evaluation.tex
\subsection{Quantitative Evaluation}
\label{sec:eval}

%Our \dataset is challenging for previous models, but \treenet gives visually better
%results (\cf Fig.~\ref{fig:prevresults} vs. Fig.~\ref{fig:prevresults}).
In this
section, we perform a rigorous evaluation to quantitatively verify the qualitative
observation that \treenet synthesises more realistically looking trees compared to state-of-the-art methods. %  are closer to training samples.
We show that both machines and humans mistake \treenet samples more often for training data compared to the other methods.

In the first
two experiments, we train a CNN classifier to distinguish between \real samples
and those generated by different models.
%These show how often a computer mistook
%\fake samples from each generative model as \real. To this end
In both cases, we used the LeNet~\cite{bottou1994comparison} architecture as the CNN of choice; the reason being that we do
not want a very deep classifier network so as to avoid overfitting. In the last
experiment, we asked human participants to mark samples as \real or \fake. For
each experiment, we report the false positive rate which is the percentage of
\fake samples mistaken for \real (\cf Table~\ref{tab:eval}).

\begin{table}[t]
\caption{False positive rates of generative models under different
    experiments indicating how often \fake samples from
    each method were mistaken for \real. \treenet results were consistently most
    indistinguishable from \real samples than other methods.}
    \label{tab:eval}
     \centering
    \begin{small}
    \begin{tabular}{lcccc}
        \toprule
                         & \dcgan \cite{Radford:2015:dcgan}
                         & \vaegan \cite{Larsen:2015:vaegan}
                         & \draw \cite{Gregor:2015:draw}
                         & \textbf{\treenet} \\ \midrule
        Artefacts study  &  6.36 &  9.87 &  9.65 & \textbf{46.61} \\
        Realism study    &  4.92 &  3.03 &  4.38 & \textbf{50.24} \\
        User study       & 33.60 & 33.17 & 39.97 & \textbf{52.11} \\ \bottomrule
    \end{tabular}
    \end{small}
\end{table}

%\PAR{\allinone evaluation using LeNet classifier}
\PAR{Quantifying visual artefacts}
In the first experiment, we quantitatively show that \treenet produces less obvious visual artefacts compared to the other approaches.
%This experiment tests the realism of images generated by all generative models in
%a joint fashion and directly compares their quality.
We created a training dataset
of 50\% \real samples and 50\% \fake samples selected equally and randomly from
the results of \dcgan, \vaegan, \draw and \treenet. We trained LeNet to classify
samples as \real or \fake using these labelled images in mini-batches of size
$64$, learning rate $0.0001$, for $200$ epochs, such that each epoch corresponds
to a full training set pass. The classifier stabilised after training on $3000$
labelled samples, learning to recognise the artefacts produced by all models
jointly. We tested previously unseen samples of each generative model and report
their respective false positive rates. \treenet results were mistaken for \real
almost 50\% of the times, compared to less than 20\% for previous models
(\cf Table~\ref{tab:eval}, first row).
This shows that other methods create visual artefacts that are easier to identify for a CNN. In other words, this shows that \treenet \emph{produces less obvious
visual artefacts than previous models}.

%\PAR{\onevsall evaluation using LeNet classifier}
\PAR{Quantifying realism}
The previous experiment demonstrated that \treenet generates more realistic samples compared to other approaches.
However, it does not allow us to measure how realistic these samples are.
The second experiment aims to tests the realism of the synthesised images by allowing the classifier to focus on the specific artefacts generated by each model.
%generative model in isolation from the other models.
For each model, we trained a separate classifier by training on an equal number of
\real and \fake images generated by the same model. Thus, the classifier
for each model learned to recognise the artefacts produced by the corresponding
model. The hyperparameters are the same as for the previous experiment. % \allinone evaluation.
Table~\ref{tab:eval} (second row) again shows %a higher false positive rate for \treenet compared to other methods.
%The high false
%positive rate of \treenet (\cf Table~\ref{tab:eval})
%This shows that even if the classifier is able to focus on the specific artefacts generated by \treenet, its
\treenet artefacts
are \emph{visually less objectionable than those of other methods}.

%Both experiments are necessary for completeness. %\allinone
%The first compares
%all models on the same trained classifier, but it is possible that the classifier
%might be confused because different methods produce different artifacts.
%%\onevsall
%The second experiment allows the classifier to focus on artifacts of each model in isolation,
%but is not a direct comparison between models because they are all tested on
%separately trained classifiers. \treenet gives high false positive rates for
%both experiments, demonstrating its efficacy over previous approaches. % making a foolproof case for its efficacy.

\PAR{Perceptual evaluation}
Complimentary to a CNN classifier, we also used human perception to rate the
quality of different generative models. We performed an anonymous user study,
where we asked participants to judge if a given tree sample is \real or \fake.
We conducted the study with 26 participants rating a total of 1590
images. The false positive rate shows that \treenet samples were mistaken for
\real samples more frequently than those of the other models (\cf Table~\ref{tab:eval}, third row).
Participants also provided subjective feedback that branch connectivity and
non-branch-like shapes \eg, blobs were the primary criteria for discerning
samples. This study is a useful addition to CNN-based evaluation because
humans tend to observe global features while CNNs may be biased by local
statistics. Please see the supplemental material for details, screenshots and
subjective feedback of the user study.

% \begin{figure*}
%     \begin{subfigure}[t]{0.32\textwidth}
%         \includegraphics[width=\columnwidth]{figures/evaluation/all_in_one_evaluation_snapshot.pdf}
%         \caption{\allinone classifier evaluation}
%         \label{fig:all-in-one}
%     \end{subfigure}
%     \begin{subfigure}[t]{0.32\textwidth}
%         \includegraphics[width=\columnwidth]{figures/evaluation/one_against_all_evaluation.pdf}
%         \caption{\onevsall classifier evaluation}
%         \label{fig:one-against-all}
%     \end{subfigure}
%     \begin{subfigure}[t]{0.32\textwidth}
%         \includegraphics[width=\columnwidth]{figures/evaluation/user_study.pdf}
%         \caption{Perceptual evaluation via user study}
%         \label{fig:userstudy}
%     \end{subfigure}
%     \caption{False positive rates (mean and variance) from different evaluation
%         protocols show the percentage of \fake samples mistaken as \real.
%         \treenet results are classified \real more frequently that \dcgan,
%         \vaegan and \draw.}
%     \label{fig:eval}
% \end{figure*}

%% file: sections/conclusion.tex
\section{Conclusion}
In this paper, we have presented \dataset, a novel dataset for benchmarking the ability of generative models to synthesise fine details.
We have demonstrated that this task is very challenging for existing generative models. We have propose \treenet, a network architecture able to better  synthesise fine details and avoid visual artefacts, and shown that it produces more realistic results through extensive qualitative and quantitative experiments.

% \treenet produces grayscale images even though the input is binary. This can
% skew evaluation because classifier CNNs may base their decisions on low level
% image statistics like grayscale values instead of overall structure. We
% resolved this by a binarisation post-process.
% Yet, the ideal solution would be to directly generate
% binary output. % from the decoder.

Table \ref{tab:eval} %Quantitative evaluation (\cf Table \ref{tab:eval})
suggests that \treenet
almost achieves the theoretically optimal false positive rate of 50\%. % false positive rate.
At the same time, \treenet still produces some residual artefacts like minor blur,
even though significant less than previous methods.
While our quantitative experiments thus demonstrate that \treenet performs better \emph{relative} to existing approaches, they should not be understood as \emph{absolute} measurements.
Choosing a network architecture for measuring the absolute false positive rate is still very much an open problem.
Still, we believe that our \dataset dataset will be helpful when tackling the problem of  evaluating the performance of generative models in a self-supervised fashion.
%The high false positive rate thus is part is
%This shows that there is
%scope for better networks for 2D sketching, and better evaluation protocol.
%We believe that our dataset will useful towards these goals.

%\para{Limitations}

%\para{Future work}

%\para{Conclusion}
%We demonstrated a neural network to learn and synthesise fine-grained 2D sketching
%and presented a dataset of tree skeletons to serve as a benchmark for future
%generative models. We hope that our work will inspire research on highly detailed
%sketching, and also open the doors for deep learning in vegetation modelling.

%% file: sketchgen-supp.tex
\section{Supplementary material}

This supplementary material provides additional details on our new \dataset dataset, additional qualitative results, and details on our quantitative experiments.
These details were left out of the paper due to space constraints.

The remainder of the supplementary material is structured as follows:
Sec.~\ref{training-data} provides additional details on our new dataset.
Sec.~\ref{sec:qualitative} shows additional qualitative results.
Sec.~\ref{sec:loss} analyses the behaviour of our loss functions during training.
%Sec.~\ref{sec:ablation} extends the ablation study presented in Fig.~6 in the paper.
Sec.~\ref{sec:eval_details} describes details of the setup used for the quantitative experiments presented in Sec.~5.2 in the paper.
Finally, Sec.~\ref{sec:user} provides details on the user study we performed to perceptually compare our \treenet method to existing state-of-the-art generative models.

\subsection{Training data}\label{training-data}

\begin{figure*}
    \centering
    \includegraphics[width=0.62\textwidth]{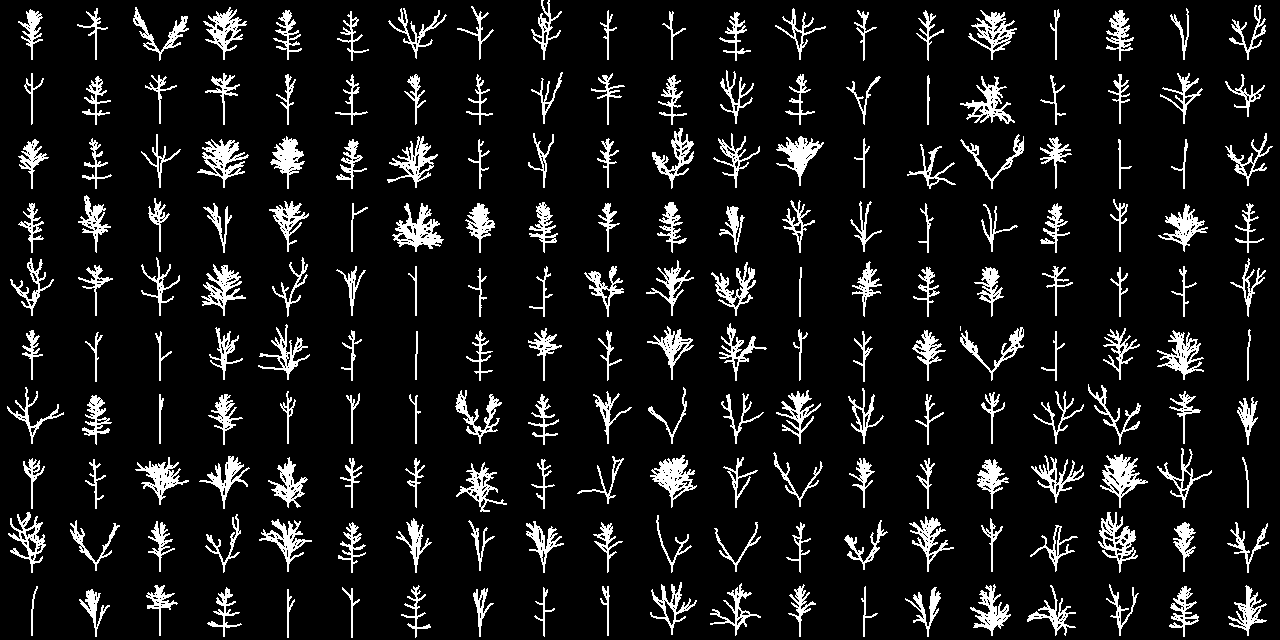} \\
    \includegraphics[width=0.62\textwidth]{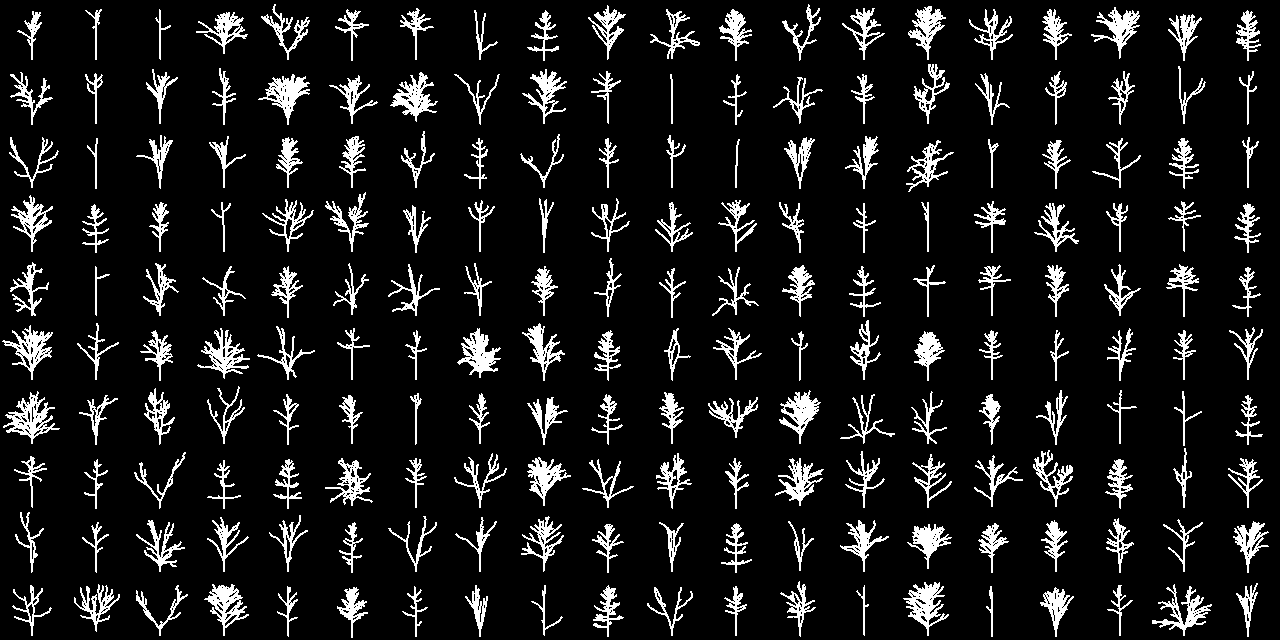} \\
    \includegraphics[width=0.62\textwidth]{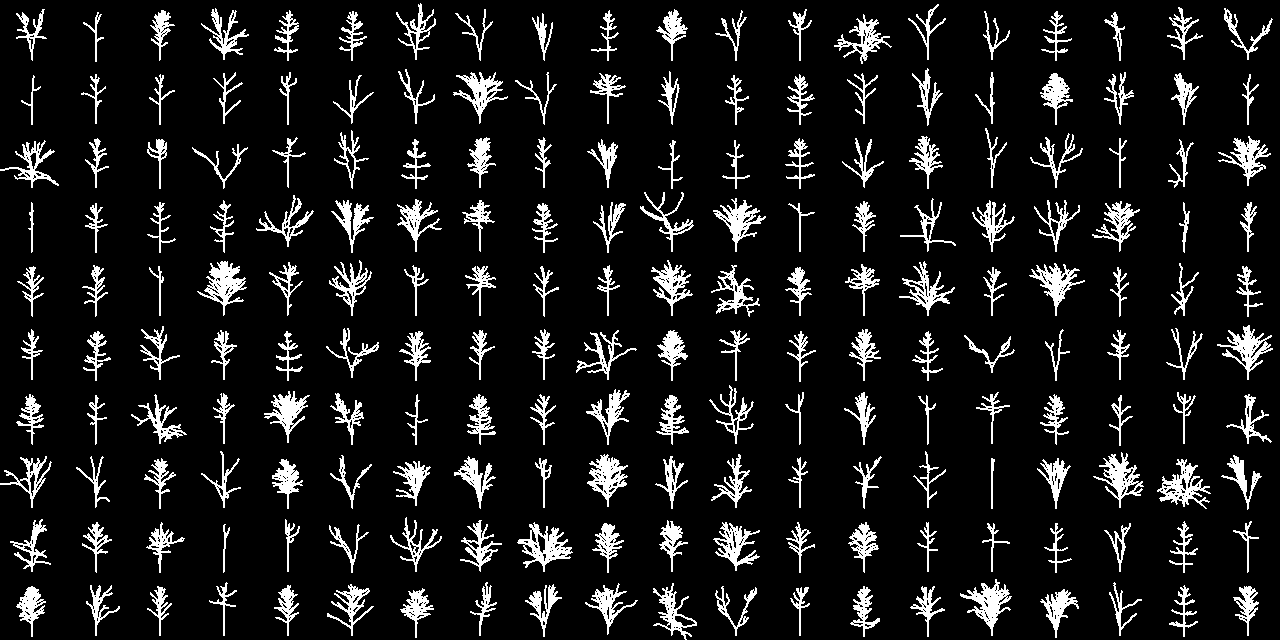} \\
    \includegraphics[width=0.62\textwidth]{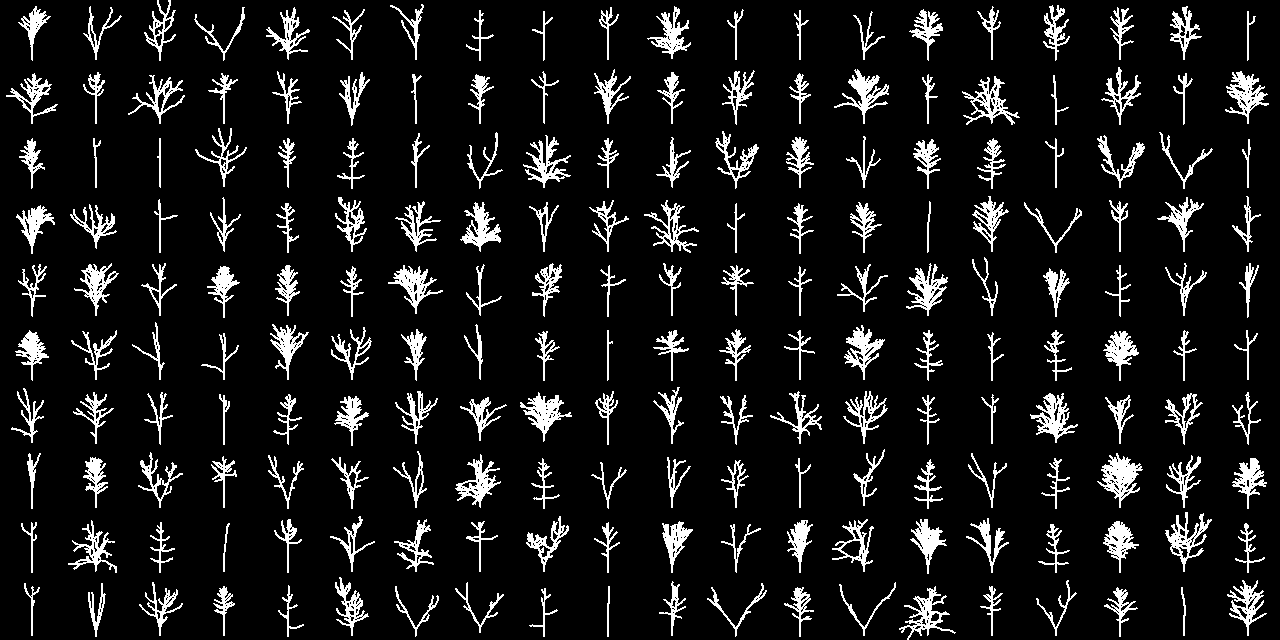}
    \caption{A visualisation of 800 examples out of the 15000 training samples contained in the proposed \dataset dataset.}
    \label{fig:supp:dataset}
\end{figure*}

\begin{figure*}
    \centering
    \begin{subfigure}{0.19\textwidth}
        \includegraphics[width=\columnwidth]{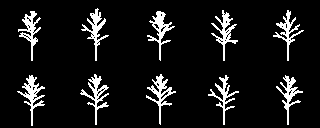}
        \caption{Small maple}
    \end{subfigure}
    \begin{subfigure}{0.19\textwidth}
        \includegraphics[width=\columnwidth]{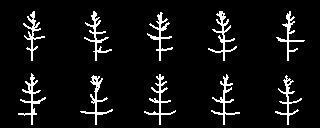}
        \caption{Pine}
    \end{subfigure}
    \begin{subfigure}{0.19\textwidth}
        \includegraphics[width=\columnwidth]{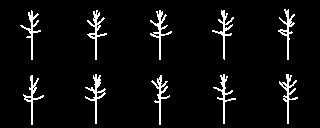}
        \caption{Popple}
    \end{subfigure}
    \begin{subfigure}{0.19\textwidth}
        \includegraphics[width=\columnwidth]{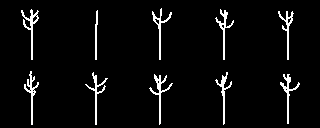}
        \caption{Kauri}
    \end{subfigure}
    \begin{subfigure}{0.19\textwidth}
        \includegraphics[width=\columnwidth]{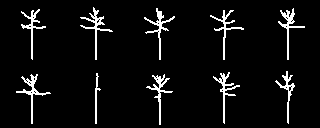}
        \caption{Teak}
    \end{subfigure}
    \begin{subfigure}{0.19\textwidth}
        \includegraphics[width=\columnwidth]{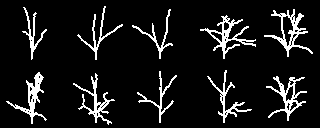}
        \caption{Chestnut}
    \end{subfigure}
    \begin{subfigure}{0.19\textwidth}
        \includegraphics[width=\columnwidth]{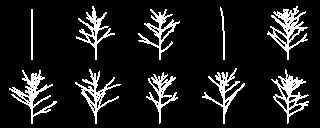}
        \caption{Beech}
    \end{subfigure}
    \begin{subfigure}{0.19\textwidth}
        \includegraphics[width=\columnwidth]{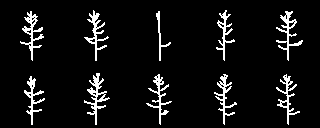}
        \caption{Larch}
    \end{subfigure}
    \begin{subfigure}{0.19\textwidth}
        \includegraphics[width=\columnwidth]{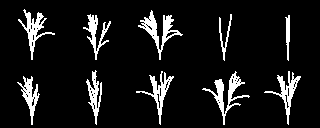}
        \caption{Callistemon}
    \end{subfigure}
    \begin{subfigure}{0.19\textwidth}
        \includegraphics[width=\columnwidth]{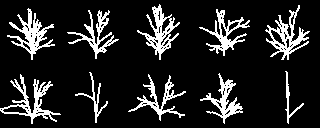}
        \caption{Linden}
    \end{subfigure}
    \begin{subfigure}{0.19\textwidth}
        \includegraphics[width=\columnwidth]{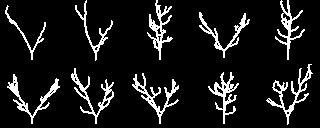}
        \caption{Acacia}
    \end{subfigure}
    \begin{subfigure}{0.19\textwidth}
        \includegraphics[width=\columnwidth]{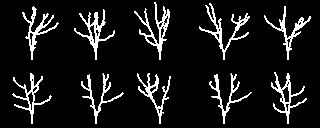}
        \caption{Cedar}
    \end{subfigure}
    \begin{subfigure}{0.19\textwidth}
        \includegraphics[width=\columnwidth]{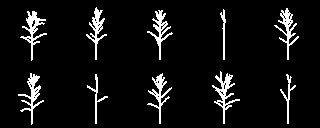}
        \caption{White birch}
    \end{subfigure}
    \begin{subfigure}{0.19\textwidth}
        \includegraphics[width=\columnwidth]{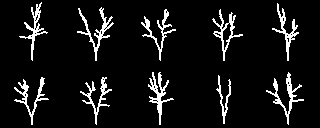}
        \caption{Jap. maple}
    \end{subfigure}
    \begin{subfigure}{0.19\textwidth}
        \includegraphics[width=\columnwidth]{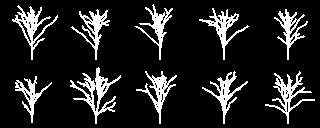}
        \caption{Elm}
    \end{subfigure}
    \caption{Examples for the different tree species in contained in the proposed \dataset dataset.}
    \label{fig:supp:species}
\end{figure*}

We will make the proposed \dataset dataset publicly available upon publication of the paper. %will be made available in machine readable binary
%HDF5 format.
In order to provide a visual overview over the dataset, Fig.~\ref{fig:supp:dataset} shows example trees from the training data.
Fig.~\ref{fig:supp:species} shows additional examples % in
%human readable form (\cf Fig.~\ref{fig:supp:dataset}) and
sorted by tree species. % (\cf Fig.~\ref{fig:supp:species}).

\PAR{Dataset details}
The dataset consists of 15000 sample tree skeleton images. There are 15
species with 1000 samples per species. For each species, we asked an artist to
create a tree model with \href{https://www.blender.org/}{Blender} using the
\href{https://wiki.blender.org/index.php/Extensions:2.6/Py/Scripts/Curve/Sapling_Tree}{Sapling
addon}.
These models use a parametric representation.
We created 250 different 3D tree
models for each species by adding %and render them from 4 random viewpoints. The 3D
%models are created in . We created a single model for each species manually and added
random jitter to the parameters selected by the artist. %to create other models.
Rendering each of the 250 models from 4 random viewpoints then resulted in the 1000 samples per species.

\begin{figure*}
    \begin{subfigure}{\textwidth}
        \centering
        \includegraphics[width=\columnwidth]{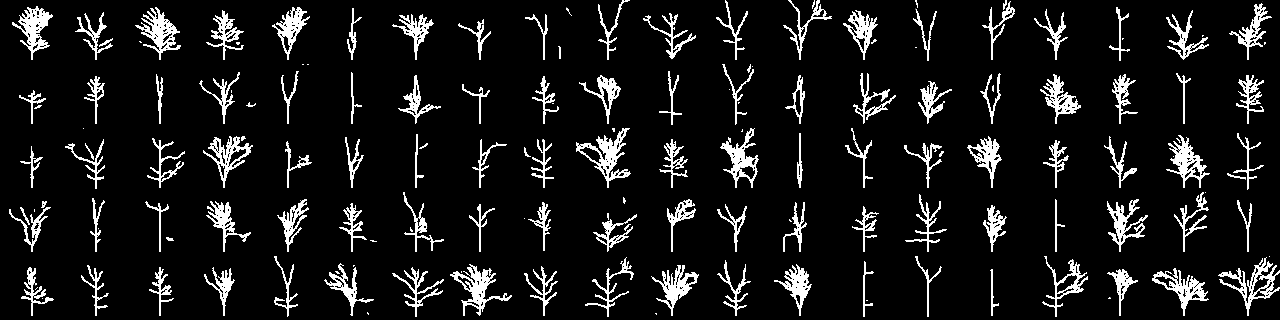}
        \caption{\dcgan \cite{Radford:2015:dcgan}}
    \end{subfigure}
    \begin{subfigure}{\textwidth}
        \centering
        \includegraphics[width=\columnwidth]{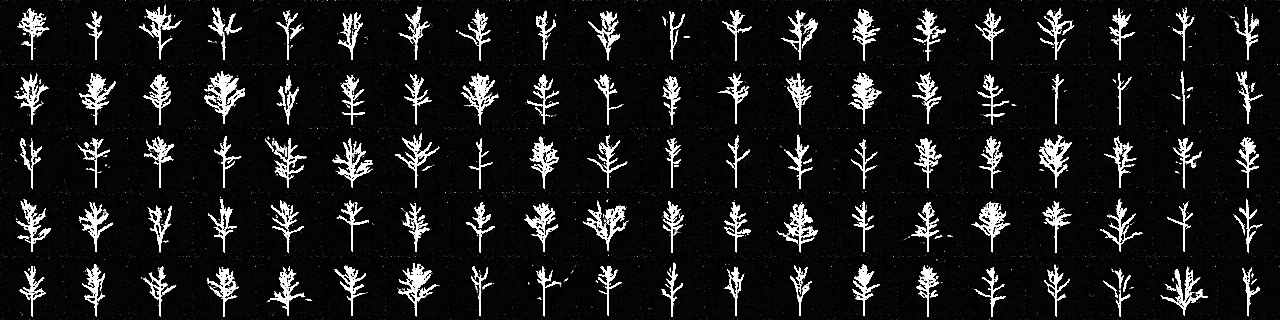}
        \caption{\vaegan \cite{Larsen:2015:vaegan}}
    \end{subfigure}
    \begin{subfigure}{\textwidth}
        \centering
        \includegraphics[width=\columnwidth]{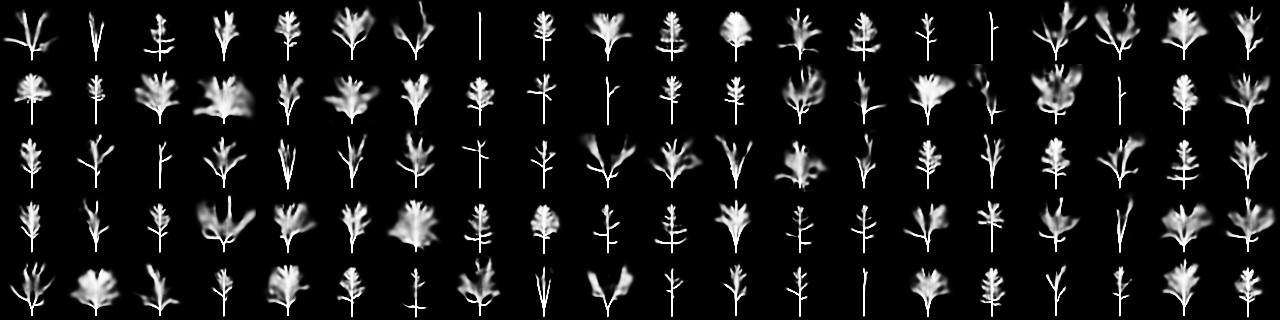}
        \caption{\draw \cite{Gregor:2015:draw}}
    \end{subfigure}
    \caption{Results produced by state-of-the-art generative models on our \dataset dataset. Noice the different artefacts (disconnected structures, high-frequency noise, and blur) of the different approaches.}
    \label{fig:supp:prev}
\end{figure*}

% ------------------------------------------------------------------------------

\subsection{Additional Qualitative Results}
\label{sec:qualitative}

This section provides additional qualitative results for both our proposed \treenet approach and state-of-the-art generative models.

In addition to Figures 1 and 3 in the paper, Fig.~\ref{fig:supp:prev} shows trees synthesised using previous generative models. % on \dataset in
%Fig.~\ref{fig:supp:prev}.
As detailed in the paper, the generated images exhibit different artifacts (disconnected and unnatural branches for \dcgan, disconnected branches and high-frequency noise for \vaegan, and blurred images for \draw).

\begin{figure}[h]
    \begin{subfigure}{0.09\columnwidth}
        \centering
        \includegraphics[width=\columnwidth]{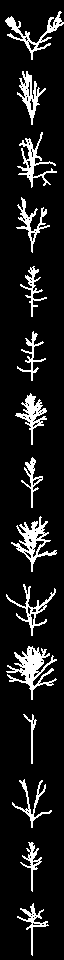}
    \end{subfigure}
    \hfill
    \begin{subfigure}{0.9\columnwidth}
        \centering
        \includegraphics[width=\columnwidth]{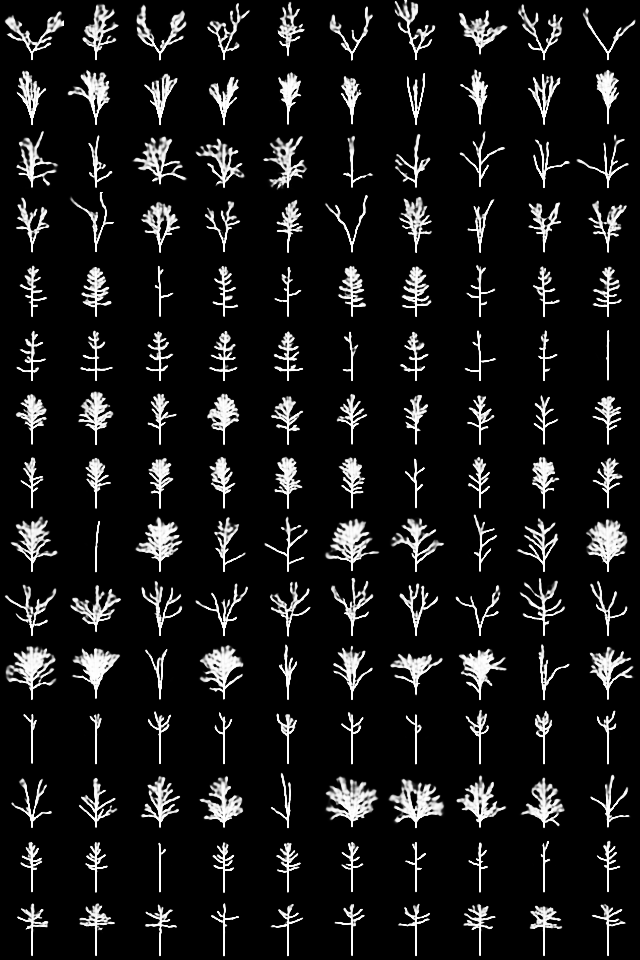}
    \end{subfigure}
    \caption{Species specific synthesis produced by the proposed \treenet method. Training sample of a species (left)
    and samples synthesised (right) by adding small perturbation to the latent
    space encoding of the training sample. The results appear to be same
    species, which suggests that species may be clustered in latent space.}
    \label{fig:speciesreconstruct}
\end{figure}

Figures 8 and 9 in the paper show trees synthesised by our \treenet method from randomly sampled points in the latent space.
Fig. ~\ref{fig:speciesreconstruct} complements these results by randomly perturbing points in the latent space corresponding to training examples.
As can be seen, the generated trees seem to appear to correspond to the same species as the training image they were generated from.
This suggests that species are clustered in the latent space.

Due to its recurrent nature, our \treenet method, which is based on an RNN, sketches trees in a temporal way.
This is illustrated in the accompanying video.
%Please see accompanying video for visualisation of the RNN's temporal sketching.
%We also show species specific synthesis (\cf Fig. ~\ref{fig:speciesreconstruct}).

% ------------------------------------------------------------------------------

\subsection{Prior Methods Implementation Details and Training Parameters}
\label{sec:methods}
\PAR{DCGAN} Our \dcgan implementation is based on the original paper's \cite{Radford:2015:dcgan} proposed network\footnote{\url{https://github.com/Newmu/dcgan_code}}.
For our experiments we changed kernel sizes to suite well for the tree skeletons and adapted in-/output channels to grayscale images (see Table~\ref{tab:dcgan} for details).
For all up- and convolution layers we applied batch normalization, which stabilizes training by shifting the input data to zero mean and unit variance.
Note that in the discriminator all \emph{ReLU} activations functions are replaced by \emph{LeakyReLU}.
We trained \dcgan on the \textsc{TreeSkel15k} dataset of size $S_T$ with mini-batches of size $S_B=128$.
Using smaller batch sizes leads to unstable training behavior.
No pre-processing is performed except for scaling the binarized input images into the range of \emph{tanh} activation function [-1, 1].
All results are obtained by training with \emph{Adam} optimizer, setting the momentum term $\beta_1$ to 0.5 and the learning rate to $0.0005$
We decreased the learning rate to avoid exploding gradients in the begging of training.
For \emph{LekyReLU} activations we set the slope of the leak to 0.2 as suggested by the authors.
The training converges after approximately 180 epochs.
An epoch corresponds to $\frac{S_T}{S_B}$ gradient updates.

\begin{table}
	\centering
	\begin{tabularx}{0.9\textwidth} { X X }
		\toprule
		Discriminator & Generator \\
		\midrule
		7$\times$7, 64 conv., LeakyReLU & 4$\times$4$\times$256, fully-connected, ReLU \\
		5$\times$5, 64 conv., LeakyReLU & 3$\times$3, 256 up-conv., ReLU \\
		3$\times$3, 64 conv., LeakyReLU & 3$\times$3, 256 up-conv., ReLU \\
		128 fully-connected & 5$\times$5, 128 up-conv., ReLU \\
		1 fully-connected, sigmoid & 7$\times$7, 64 up-conv. tanh \\
		\bottomrule
	\end{tabularx}
	\caption{Our \dcgan architecture.
	We use a different kernel sizes for the discriminator and the generator network.}
	\label{tab:dcgan}
\end{table}

\PAR{VAE-GAN} We implement \vaegan based on the proposed framework \cite{Larsen:2015:vaegan}\footnote{\url{https://github.com/andersbll/autoencoding_beyond_pixels}} adapted for binarized images.
Table~\ref{tab:vae-dcgan} lists details about the encoder and modified discriminator network.
The $l^{th}$-layer introduced by \cite{Larsen:2015:vaegan} consists of a weighted sum of different deep convolutional layers.
We found that choosing the $3^{rd}$ and $4^{th}$ convolutional layers as well as the first fully-connected layer yielded the best results.
Similar to \cite{Larsen:2015:vaegan}, we update three optimizers each responsible to minimize either encoder, decoder or discriminator gradients.
The hyperparameters remained the same as in the previously presented experiment with \dcgan.
If a parameter is not explicitly listed here, it did not change.
All learning rates for the encoder, decoder and discriminator are set to 0.0001.
The training converges after approximately 450 epochs.
An epoch corresponds to $\frac{S_T}{S_B}$ gradient updates, $S_T$ denotes dataset size and $S_B$ is the mini-batch size.

\begin{table}
	\centering
	\begin{tabularx}{0.9\textwidth} { X X }
		\toprule
		Encoder & Discriminator \\
		\midrule
		5$\times$5, 64 conv., ReLU & 5$\times$4, 64 conv., LeakyReLU \\
		3$\times$3, 128 conv., ReLU & 3$\times$3, 128 conv., LeakyReLU \\
		3$\times$3, 256 conv., ReLU & 3$\times$3, 256 conv., LeakyReLU \\
		256 fully-connected & 3$\times$3, 256 conv., LeakyReLU \\
		256 fully-connected & 256 fully-connected, LeakyReLU \\
		& 50\% dropout \\
		& 1 fully-connected, sigmoid \\
		\bottomrule
	\end{tabularx}
	\caption{Our \vaegan architecture.
	Note the generator as listed in Table~\ref{tab:dcgan} becomes the decoder network.
	Encoder as well as discriminator are composed of strided 2D convolutional layers.
	Dropout is only applied to the discriminator network and acts as a regularization to prevent overfitting.}
	\label{tab:vae-dcgan}
\end{table}

\PAR{DRAW} We follow the approach described in \cite{Gregor:2015:draw}\footnote{\url{https://github.com/ericjang/draw}}.
The model architecture as well as about the used objectives can be found in \cite{Gregor:2015:draw}.
We did not change any of the building blocks of \draw and train the network as proposed with the \emph{Adam} optimizer, the momentum set to $\beta_1=0.9$.
The learning rate is set to 0.0005.
Training converges after approximately 300 epoch.
An epoch corresponds to $\frac{S_T}{S_B}$ gradient updates, $S_T$ denotes dataset size and $S_B$ is the mini-batch size.
All shown samples, we generate from a normal distribution $z \sim \mathcal{N}(0,1)$.
We take the same number of dimension for the latent space as suggested in the paper: $z \in \mathbb{R}^{100}$.
In this experiment we enabled the attention mechanism and set the read and write window size to $3\times3$ and $5\times5$ respectively.
The number of glimpses or equivalently defined as time steps, is set to 64.
For both recurrent networks, encoder and decoder, we choose the hidden vectors $h^{enc}$ and $h^{dec}$ to have dimension 256.

% ------------------------------------------------------------------------------

\subsection{Evaluation of the Loss Functions}
\label{sec:loss}
\begin{figure}
    \centering
    % \begin{subfigure}{\columnwidth}
    %     \centering
    %     \includegraphics[width=\columnwidth]{losses/prior}
    %     \caption{Prior loss}
    %     \label{fig:supp:prior}
    % \end{subfigure}
    \begin{subfigure}{\columnwidth}
        \centering
        \includegraphics[width=\columnwidth]{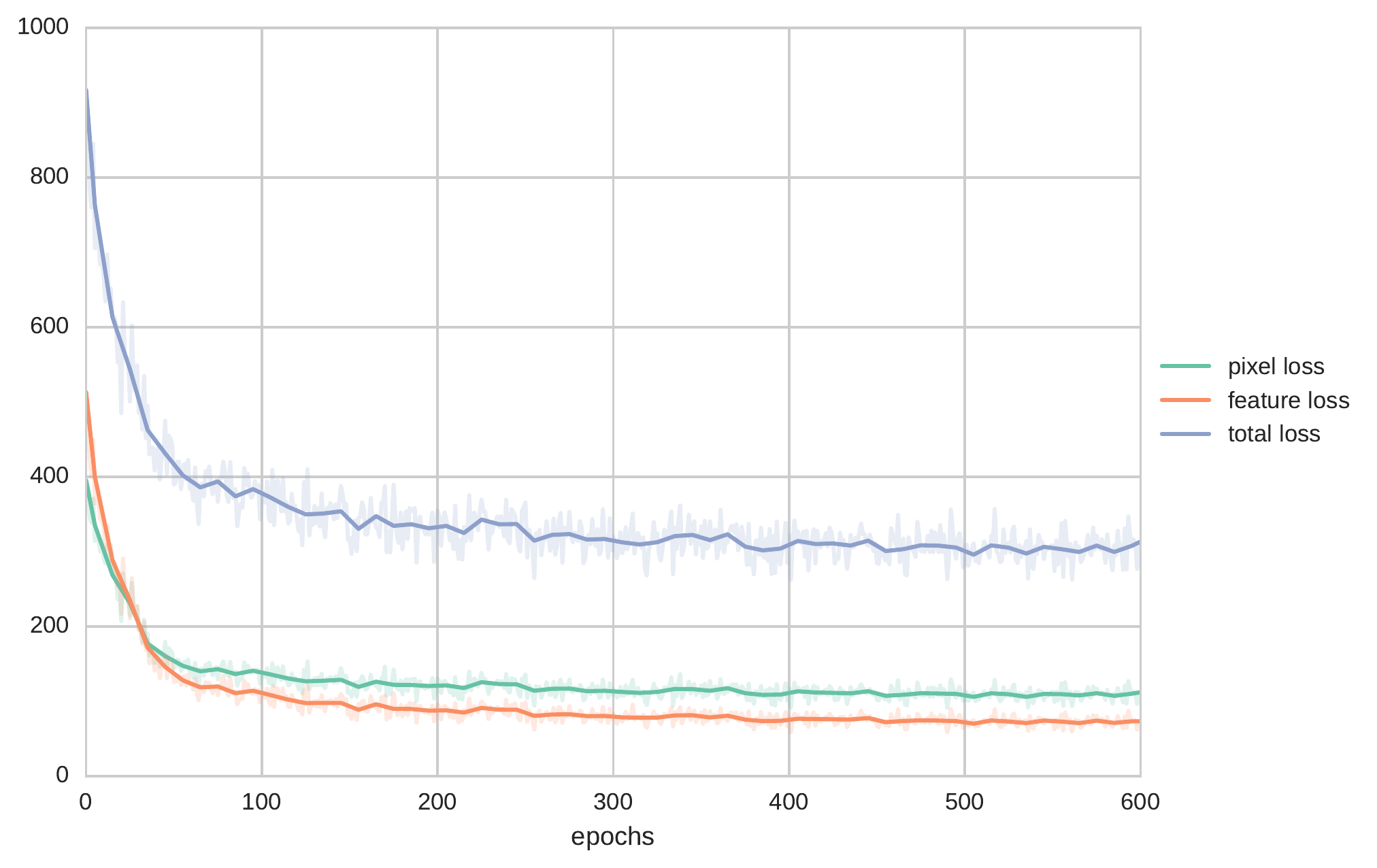}
        \caption{Pixel loss, feature loss and their weighted sum}
        \label{fig:supp:encoder}
    \end{subfigure}
    \begin{subfigure}{\columnwidth}
        \centering
        \includegraphics[width=\columnwidth]{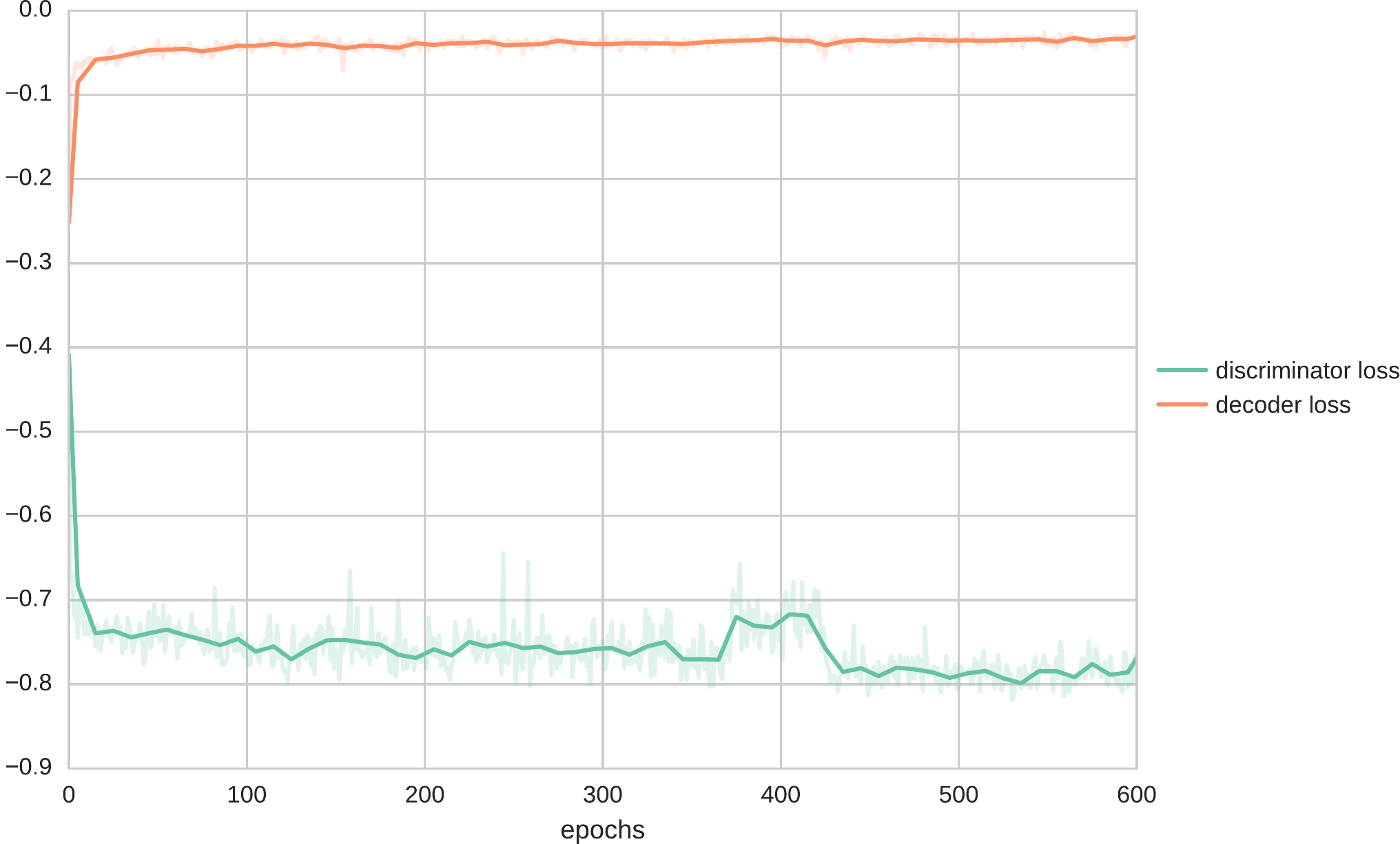}
        \caption{Discriminator and generator or decoder losses}
        \label{fig:supp:discriminator}
    \end{subfigure}
    \caption{Evolution of the different loss functions as a function of the number of training epochs.}
    \label{fig:supp:losses}
\end{figure}

Fig.~\ref{fig:supp:losses} shows the evolution of the different losses
as a function of the number of training epochs.
% The initial growth of the KL-divergence between the latent space distribution and the Gaussian
% distribution indicates that the latent space assumes a normal distribution as
% training progresses. The flattening in the end indicates convergence
% (\cf Fig.~\ref{fig:supp:prior}).
Both pixel and feature losses decay slowly as the autoencoder learns to
synthesise realistic samples (\cf Fig.~\ref{fig:supp:encoder}).
The decreasing discriminator loss indicates that the binary
classification of the discriminator improves while the increasing generator/decoder loss illustrates that the generator becomes increasingly more capable of synthesising realistic
samples (\cf Fig.~\ref{fig:supp:discriminator}).

\subsection{Evaluation Details}
\label{sec:eval_details}

In the following, we expand on the quantitative evaluation presented in Sec. 5.2 of the paper. More precisely, we analyse the impact of using more training samples to train the classifier used for quantitative evaluation.

\begin{figure}[h]
    \centering
    \includegraphics[width=\columnwidth]{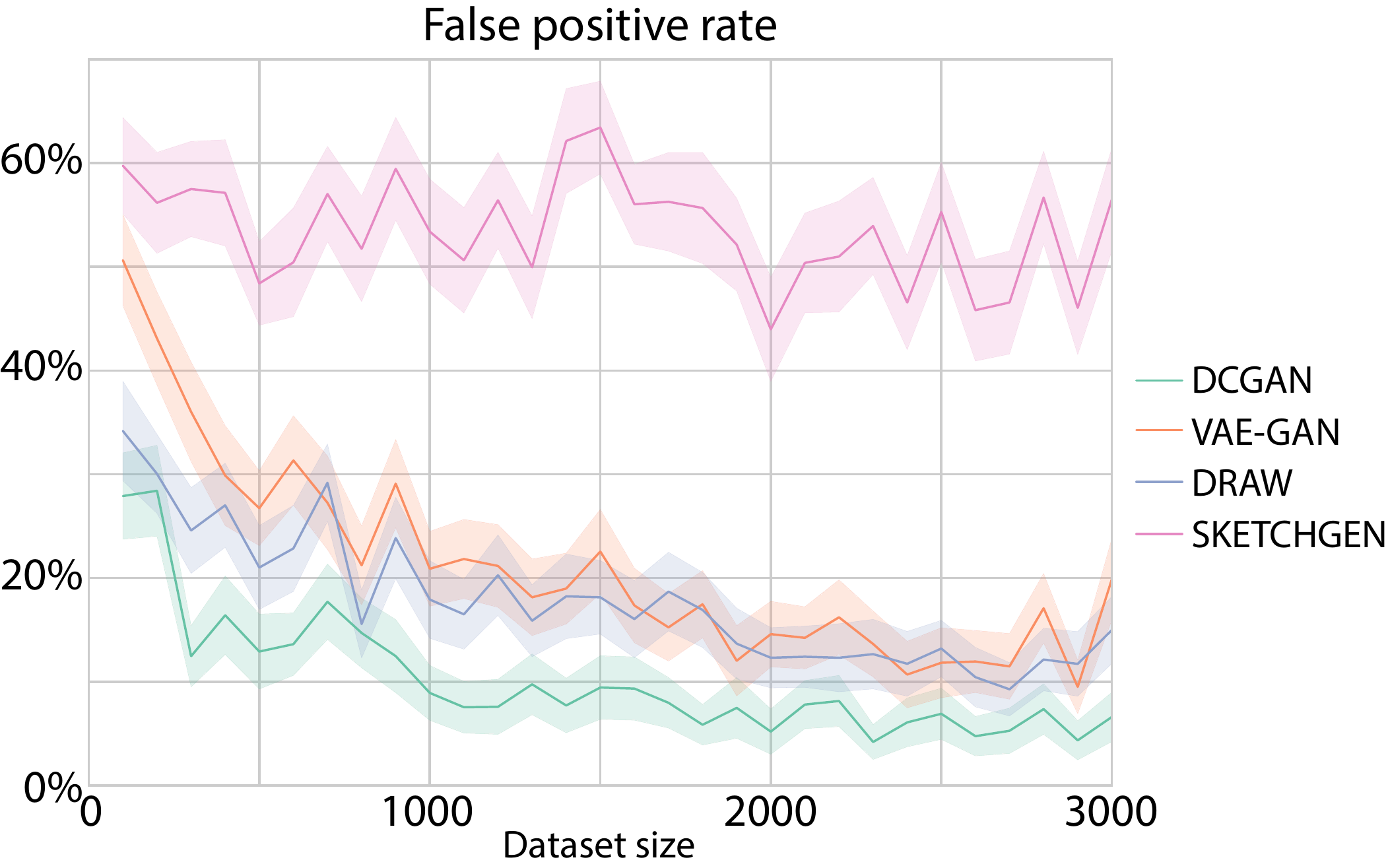}
    \caption{Effect of training dataset size on CNN classifier quality for
    the quantitative evaluation of \treenet results.}
    \label{fig:supp:evolution}
\end{figure}

\begin{figure}[h]
    \begin{subfigure}{\columnwidth}
        \includegraphics[width=\columnwidth]{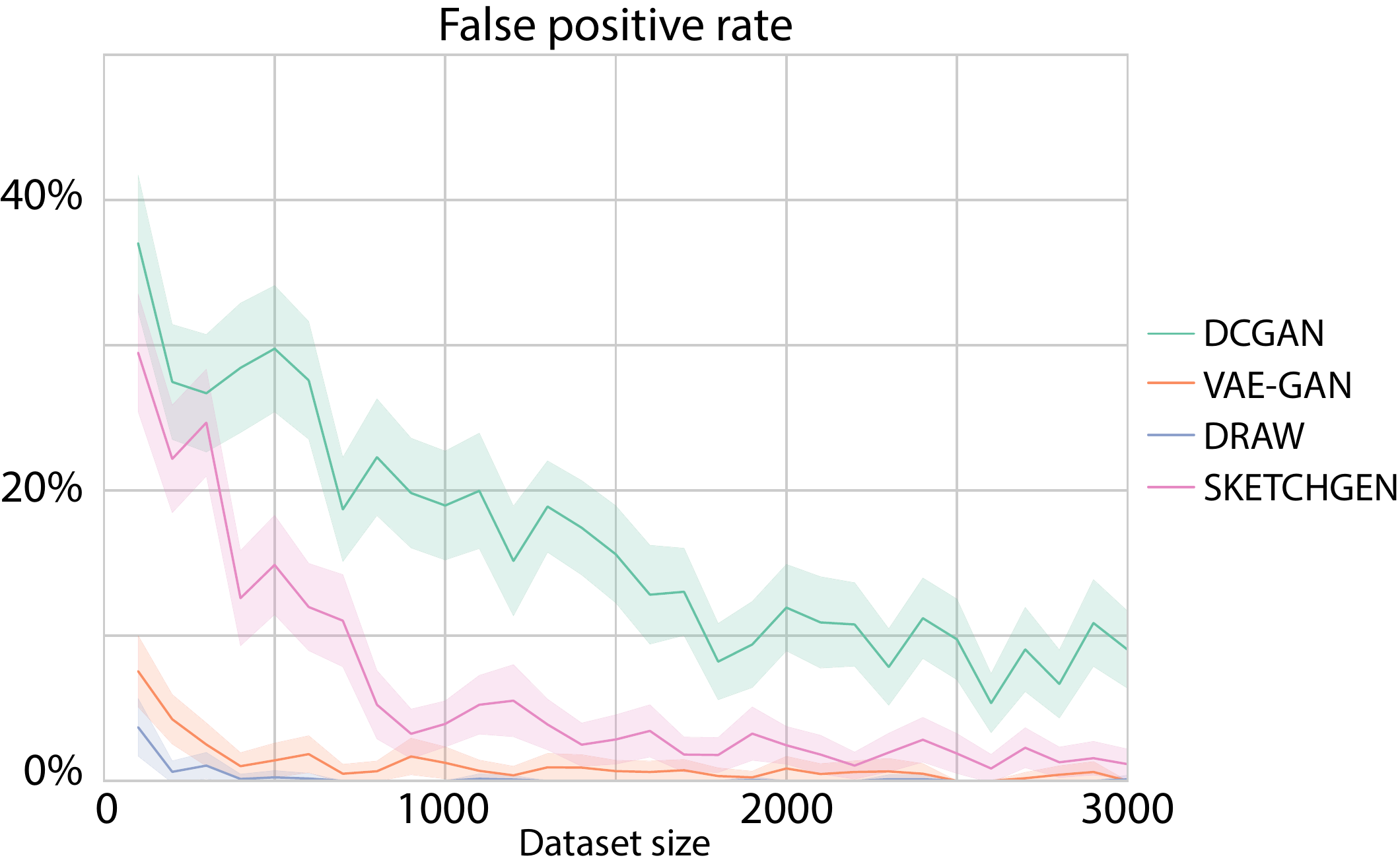}
        \caption{Evaluation on grayscale results}
        \label{fig:supp:binarised:grayscale}
    \end{subfigure}
    \begin{subfigure}{\columnwidth}
        \includegraphics[width=\columnwidth]{supplement/evaluation/all_in_one_all_models_evolution}
        \caption{Evaluation on binarised results}
         \label{fig:supp:binarised:binarised}
    \end{subfigure}
    \caption{Evaluation on binarised versus grayscale results.}
    \label{fig:supp:binarised}
\end{figure}

\PAR{Effect of training dataset size on evaluators}
A quantitative evaluation using a CNN classifier is valid only if the
classifier is well trained. The number of labelled images used for
training has a direct impact on the quality of the classifier. We visualise
classification results of the first CNN based evaluation
(titled ``Quantifying visual artefacts'' in Sec. 5.2 of the paper) as a
function of the number of images used to train the classifier (\cf Fig.~\ref{fig:supp:evolution}). We observe
that the false positive rates stabilise after training the classifier on
3000 or more samples. The quality of the evaluation also increases as
the training dataset increases and starts flattening around 3000. For
small training sizes, the classifier has the same poor false positive
rate,\ie, it is equally confused, for all methods. As the training set size grows, the
false positive rate for \treenet remains close to 50\%, while the rates drop
steeply for previous methods (\cf Fig.~\ref{fig:supp:evolution}). A similar
trend was observed for the other CNN based evaluation discussed in Sec. 5.2 in
the paper.

\PAR{Evaluation on binarised versus grayscale samples}
Even though training data for all generative models is binarised, our
model and \draw synthesise grayscale images. During evaluation,
a CNN classifier could thus only learn to distinguish between real and synthesised images on the basis on whether they
are binary or grayscale. Thus, DCGAN seems to perform best even though
it has the worst visual artefacts from a human perspective (\cf Fig.~\ref{fig:supp:binarised:grayscale}). Our goal is to evaluate high level
structure; we therefore binarise all results such that any pixel with
intensity 0.6 or higher is snapped to 1, and 0 otherwise. The CNN classifier now
learns to distinguish between high level visual content and realism
because all images are binary and low level statistics do not dominate
the classification (\cf Fig.~\ref{fig:supp:binarised:binarised}).
As a result, we used the binarised variant for the quantitative results presented in Sec. 5.2 of the paper.

\subsection{User Study Details}
\label{sec:user}
This section provides additional details on the user study we used for a perceptual evaluation of the different generative model (\cf Sec.~5.2 in the paper).
\def \wr {0.6}
\begin{figure}
    \centering
    \begin{subfigure}{\columnwidth}
        \centering
        \fbox{\includegraphics[width=\wr\columnwidth]{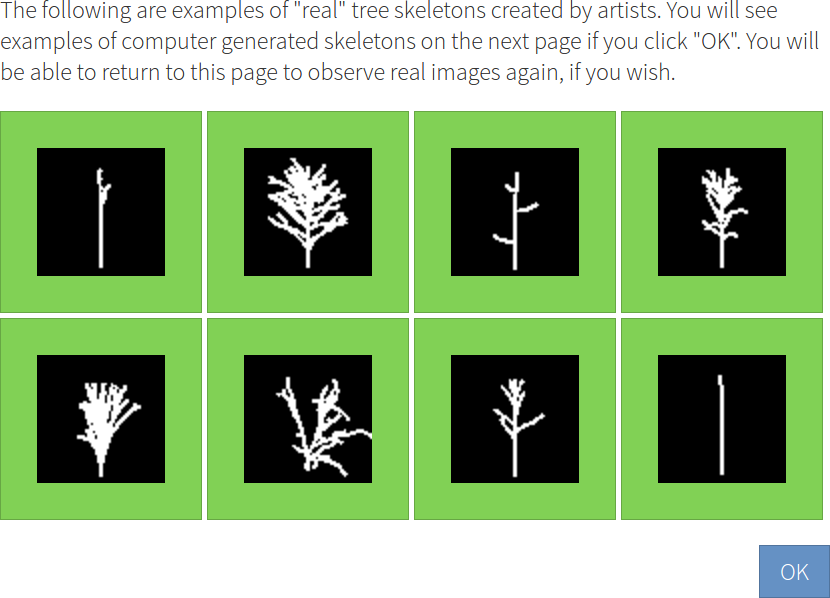}}
        \caption{Examples of \real samples shown to participants}
    \end{subfigure}
    \begin{subfigure}{\columnwidth}
        \centering
        \fbox{\includegraphics[width=\wr\columnwidth]{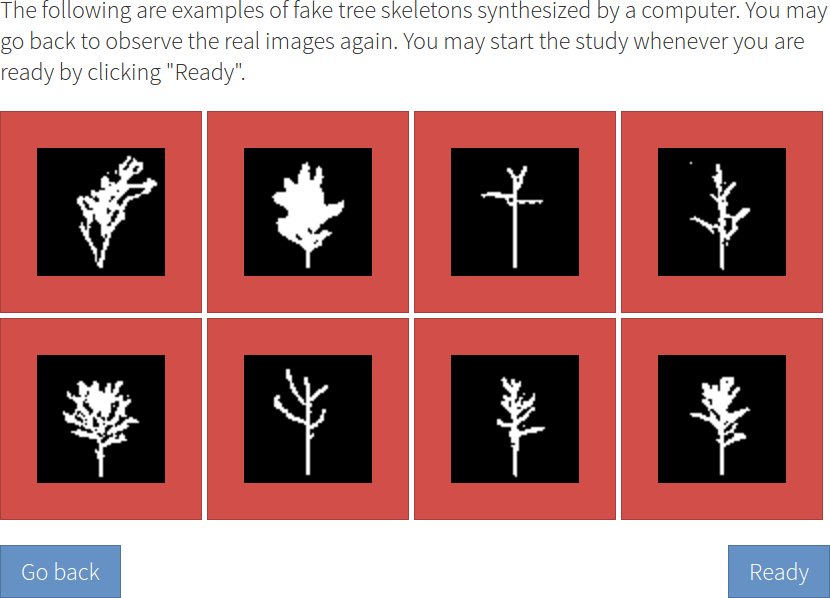}}
        \caption{Examples of \fake samples shown to participants}
    \end{subfigure}
    \begin{subfigure}{\columnwidth}
        \centering
        \fbox{\includegraphics[width=\wr\columnwidth]{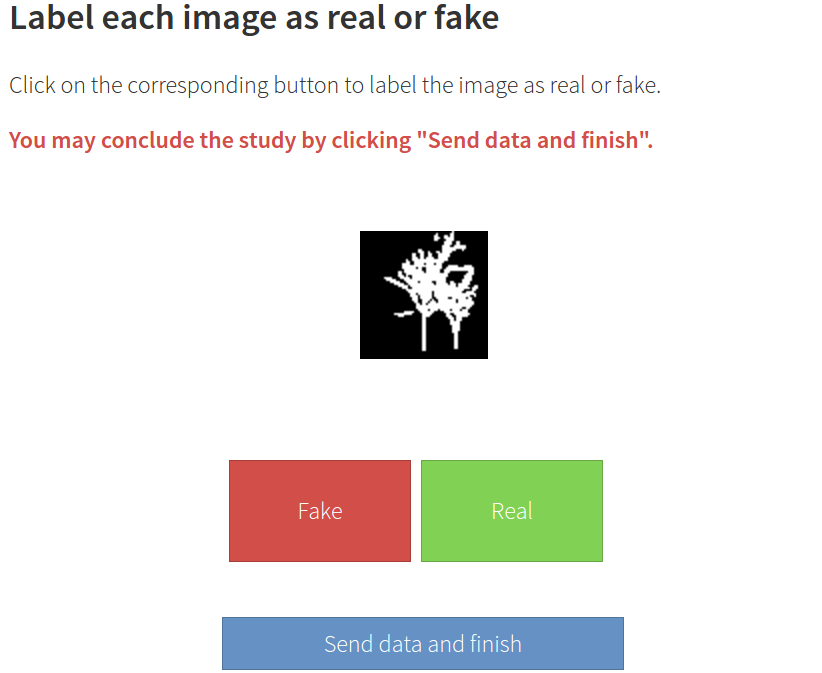}}
        \caption{Classification task}
    \end{subfigure}
    \caption{Screenshots from our user study.}
    \label{fig:supp:shots}
\end{figure}

\PAR{User study setup} We first showed participants 8 examples of \real samples (called
\emph{real} in the study) drawn randomly from the
\dataset dataset, followed by 8 examples of \fake samples (called \emph{fake} in the study)
drawn randomly from the results of \dcgan, \vaegan, \draw and \treenet
(\cf Fig.~\ref{fig:supp:shots}).
Participants were not allowed to return to the 8 \emph{real} and
\emph{fake} samples shown at the beginning once they began the
classification task. We did not enforce a minimum or maximum on the
number of samples each participant had to classify. We upsampled all
images to 128$\times$128 to aid legibility in the study.

\PAR{Subjective feedback}
We also asked participants to describe the criteria they used to
distinguish between samples. Some of the responses were:
\begin{myitemize}
\item
  \emph{not connected white tree parts, not normal shapes of the tree}
\item
  \emph{symmetry, discontinuous branches, orientation of branches,
  random pixels}
\item
  \emph{disconnected components, thick blobs}
\item
  \emph{fake ones seem to have disconnected far components than the main
  trunk, and also sometimes unrealistic shapes}
\item
  \emph{too large white blobs. White parts not attached to tree}
\end{myitemize}

These indicate that participants relied on branch connectivity and
shapes that do not correspond to realistic branches, \eg, blobs, to distinguish
between the samples. This shows that the user study is a useful addition
to our CNN-based evaluation because humans tend to observe global features
while CNNs may be biased by local image statistics.